\documentclass[twoside,11pt]{article}

\usepackage{jmlr2e}
\usepackage{url}
\usepackage{color}
\usepackage{amsmath}
\usepackage{bm}
\usepackage{float}
\usepackage{url}
\usepackage{algorithm}
\usepackage{algorithmic}
\usepackage{graphics}
\usepackage{subfigure}
\usepackage[small]{caption}
\usepackage{bbm}
\usepackage{amsfonts}
\usepackage{amssymb}
\usepackage{graphicx}

\newcommand{\be}{\begin{equation}}
\newcommand{\ee}{\end{equation}}
\newcommand{\bi}{\begin{itemize}}
\newcommand{\ei}{\end{itemize}}
\newcommand{\bea}{\begin{eqnarray}}
\newcommand{\eea}{\end{eqnarray}}

\newcommand{\bfalpha}{\boldsymbol{\alpha}}

\newcommand{\bflambda}{\boldsymbol{\lambda}}

\newcommand{\bftheta}{\boldsymbol{\theta}}

\newcommand{\bff}{\mathbf{f}}

\newcommand{\bfu}{\mathbf{u}}

\newcommand{\bfx}{\mathbf{x}}
\newcommand{\bfy}{\mathbf{y}}
\newcommand{\bfz}{\mathbf{z}}

\newcommand{\bfzero}{\mathbf{0}}

\newcommand{\cut}[1]{}

\newcommand{\dif}{\textrm{d}}
\newcommand{\veC}{\textbf{\hspace{-0.001in}:}}
\newcommand{\kMatrix}{\mathbf{K}}

 \DeclareMathOperator{\tr}{tr}

\DeclareMathOperator{\vecO}{vec} 

\providecommand{\abs}[1]{\lvert#1\rvert}

\firstpageno{1}

\title{Variational Inducing Kernels for Sparse Convolved Multiple Output Gaussian Processes}

\author{\name Mauricio A. \'{A}lvarez \email alvarezm@cs.man.ac.uk\\
\addr School of Computer Science\\
       University of Manchester\\
       Manchester, UK, M13 9PL\\
\AND
\name David Luengo \email luengod@ieee.org\\
\addr Dep. Teor\'ia de Se\~nal y Comunicaciones\\
Universidad Carlos III de Madrid\\
28911 Legan\'es, Spain\\
\AND
\name Michalis K. Titsias \email mtitsias@cs.man.ac.uk\\
\name Neil D. Lawrence \email neill@cs.man.ac.uk \\
       \addr School of Computer Science\\
       University of Manchester\\
       Manchester, UK, M13 9PL}

\begin{document}

\maketitle

\begin{abstract}
 Interest in multioutput kernel methods is increasing, whether under
  the guise of multitask learning, multisensor networks or structured
  output data. From the Gaussian process perspective a multioutput
  Mercer kernel is a covariance function over correlated output
  functions. One way of constructing such kernels is based on convolution
  processes (CP). A key problem for this approach is efficient
  inference. \citet{Alvarez:sparse2009} recently presented a
  sparse approximation for CPs that enabled efficient inference.  In
  this paper, we extend this work in two directions: we introduce the
  concept of variational inducing functions to handle potential
  non-smooth functions involved in the kernel CP construction and we
  consider an alternative approach to approximate inference based on
  variational methods, extending the work by
  \citet{Titsias:variational09} to the multiple output case. We
  demonstrate our approaches on prediction of school marks, compiler
  performance and financial time series.

\end{abstract}


\section{Introduction}
\label{Sec:Introduction}
Gaussian processes (GPs) are flexible non-parametric models which
allow us to specify prior distributions and perform inference of
functions. A limiting characteristic of GPs is the fact that the
computational cost of inference is in general $O(N^3)$, $N$ being the
number of data points, with an associated storage requirement of
$O(N^2)$. In recent years a lot of progress
\citep{Csato:sparse00,Lawrence:ivm02,Seeger:fast03,Snelson:pseudo05,Quinonero:unifying05}
has been made with approximations that allow inference in $O(K^2N)$
(and associated storage of $O(KN)$, where $K$ is a user specified
number. This has made GPs practical for a range of larger scale
inference problems. 
\\
In this paper we are specifically interested in developing priors over multiple functions. While such priors can be
trivially specified by considering the functions to be independent,
our focus is on priors which specify correlations between the
functions. Most attempts to apply such priors
so far \citep{Teh:semiparametric05,Rogers:towards08,Bonilla:multi07} 
have focussed on what is known in the geostatistics community as
``linear model of coregionalization'' (LMC)
\citep{Journel:miningBook78,Goovaerts:book97}. In these models the different outputs
are assumed to be linear combinations of a set of one or more ``latent
functions'' so that the $d$th output of the function,
$f_d\left(\mathbf{x}\right)$ is given by
\begin{align}\label{eq:lmc:model}
f_d\left(\mathbf{x}\right) = \sum_{q=1}^Q a_{d,q} u_q\left(\mathbf{x}\right),
\end{align}
where $u_q\left(\mathbf{x}\right)$ is one of $Q$ latent functions that,
weighted by $\left\{a_{d,q}\right\}_{q=1}^Q$, sum to form each of the
$D$ outputs. 
GP priors are placed, independently, over each of the latent functions
inducing a correlated covariance function over
$\left\{f_d\left(\mathbf{x}\right)\right\}_{d=1}^D$. Approaches to
multi-task learning arising in the kernel community \citep[see for
example][]{Evgeniou:multitask05} can also be seen to be instances of
the LMC framework.

We wish to go beyond the LMC framework, in particular, our focus is
\emph{convolution processes}
\citep{Higdon:convolutions02,Boyle:dependent04}. Using CPs for multi-output GPs was proposed by
\citet{Higdon:convolutions02} and introduced to the machine learning
audience by \cite{Boyle:dependent04}. Convolution processes
allow the integration of prior information from physical models, such
as ordinary differential equations, into the covariance
function. \citet{Alvarez:lfm09}, inspired by \cite{Gao:latent08}, have
demonstrated how first and second order differential equations, as
well as partial differential equations, can be accommodated in a
covariance function. Their interpretation is that the set of latent
functions are a set of \emph{latent forces}, and they term the
resulting models ``latent force models''. The covariance functions for
these models are derived through convolution processes (CPs). In the
CP framework, output functions are generated by convolving $Q$ latent
processes $\{u_q(\bfx)\}_{q=1}^{Q}$ with kernel functions,\footnote{Not
  kernels in the Mercer sense, but kernels in the normal sense.}
$G_{d,q}(\mathbf{x})$, associated to each output $d$ and latent force
$q$, so that we have
\begin{align}
  f_d\left(\mathbf{x}\right)& = \sum_{q=1}^Q \int_\mathcal{Z}
  G_{d,q}\left(\mathbf{x}-\mathbf{z}\right) u_q\left(\mathbf{z}\right)
  \mathrm{d}\mathbf{z}. \label{eq:CPsum}
\end{align}
The LMC can be seen as a particular case of the CP, in which the kernel
functions $G_{d,q}(\mathbf{x})$ correspond to scaled Dirac
$\delta$-function
$G_{d,q}\left(\mathbf{x}-\mathbf{z}\right)=a_{d,q}\delta(\mathbf{x}-\mathbf{z})$.
In latent force models the convolving kernel, $G_{d,r}(\cdot)$, is the
\emph{Green's function} associated to a particular differential
equation. 

A practical problem associated
with the CP framework is that in these models inference 
has computational complexity $O(N^3D^3)$ and storage requirements
$O(N^2D^2)$. 
Recently, \citet{Alvarez:sparse2009} introduced an efficient
approximation for inference in this multi-output GP model.  The idea
was to exploit a conditional independence assumption over the output
functions $\left\{f_d\left(\mathbf{x}\right)\right\}_{d=1}^D$ given a
finite number of observations of the latent functions
$\left\{\left\{u_q\left(\mathbf{x}_k\right)\right\}_{k=1}^K\right\}_{q=1}^Q$. This
led to approximations that were very similar in spirit to the PITC and
FITC approximations of
\citet{Snelson:pseudo05,Quinonero:unifying05}. In this paper we build
on the work of \citeauthor{Alvarez:sparse2009}. Their approximation
was inspired by the fact that if the latent functions are observed in
their entirety, the output functions \emph{are} conditionally
independent of one another (as can be seen in \eqref{eq:CPsum}). We
extend the previous work presented in \citet{Alvarez:sparse2009} in
two ways. First, a problem with the FITC and PITC approximations can
be their propensity to overfit when inducing inputs are optimized. A
solution to this problem was given in recent work by
\citet{Titsias:variational09} who provides a sparse GP approximation
that has an associated variational bound. In this paper we show how
the ideas of \citeauthor{Titsias:variational09} can be extended to the
multiple output case. Second, we notice that if the locations of the
inducing points, $\left\{\mathbf{x}_k\right\}_{k=1}^K$, are close
relative to the length scale of the latent function, the PITC
approximation will be accurate. However, if the length scale becomes
small the approximation requires very many inducing points. In the
worst case, the latent process could be white noise (as suggested by
\citet{Higdon:convolutions02} and implemented by
\citet{Boyle:dependent04}). In this case the approximation will fail
completely. We further develop the variational approximation to allow
us to work with rapidly fluctuating latent functions (including white
noise). This is achieved by augmenting the output functions with one
or more additional functions. We refer to these additional outputs as
the \emph{inducing functions}. Our variational approximation is
developed through the inducing functions. There are also smoothing
kernels associated with the inducing functions. The quality of the
variational approximation can be controlled both through these
\emph{inducing kernels} and through the number and location of the
inducing inputs.

Our approximation allows us to consider latent force models with a
larger number of states, $D$, and data points $N$. The use of inducing
kernels also allows us to extend the inducing variable approximation
of the latent force model framework to systems of \emph{stochastic
  differential equations} (SDEs).  In this paper we apply the variational
inducing kernel approximation to different real world datasets,
including a multivariate financial time series example.

A similar idea to the inducing function one introduced in this paper, was simultaneously proposed by
\citet{Lazaro:interdomain:2010}. \citet{Lazaro:interdomain:2010} introduced the concept of inducing feature to 
improve performance over the pseudo-inputs approach of \citet{Snelson:pseudo05} in sparse GP models. Our use
of inducing functions and inducing kernels is motivated by the necessity to deal with non-smooth latent functions 
in the convolution processes model of multiple outputs.   

\section{Multiple Outputs Gaussian Processes} 

Let $\bfy_d \in \mathbbm{R}^N$, where $d=1,\ldots,D$, be the observed
data associated with the output function $y_d(\mathbf{x})$. For simplicity,
we assume that all the observations associated with different outputs
are evaluated at the same inputs $\mathbf{X}$ (although this
assumption is easily relaxed).  We will often use the stacked vector
$\mathbf{y} = (\bfy_1, \ldots, \bfy_D)$ to collectively denote the
data of all the outputs.  Each observed vector $\bfy_d$ is assumed to be
obtained by adding independent Gaussian noise to a vector of function
values $\bff_d$ so that the likelihood is $p(\bfy_d|\bff_d) =
\mathcal{N}(\bfy_d|\bff_d, \sigma_d^2 I)$, where $\bff_d$ is defined via
(\ref{eq:CPsum}).  More precisely, the assumption in (\ref{eq:CPsum})
is that a function value $f_d(\bfx)$ (the noise-free version of
$y_d(\bfx)$) is generated from a common pool of $Q$ independent latent
functions $\{u_q(\bfx)\}_{q=1}^Q$, each having a covariance function
(Mercer kernel) given by $k_q\left(\bfx, \bfx^\prime\right)$.  Notice
that the outputs share the same latent functions, but they also have
their own set of parameters $(\{\bfalpha_{d,q}\}_{q=1}^Q,\sigma_d^2)$
where $\bfalpha_{d,q}$ are the parameters of the smoothing kernel
$G_{d,q}(\cdot)$. Because convolution is a linear operation, the
covariance between any pair of function values $f_d(\bfx)$ and $f_{d'}
(\bfx')$ is given by 
\begin{align*}
k_{f_d,f_{d'}}(\bfx,\bfx')=\text{Cov}[f_d(\bfx),f_{d'}(\bfx')] =
\sum_{q=1}^{Q} \int_\mathcal{Z} G_{d,q}(\bfx - \bfz) \int_\mathcal{Z}
G_{d',q}(\bfx' - \bfz') k_q(\bfz,\bfz') \mathrm{d}\bfz \mathrm{d}
\bfz'.  
\end{align*}
This covariance function is used to define a fully-coupled GP
prior $p(\bff_1,\ldots, \bff_D)$ over all the function values
associated with the different outputs. 

The joint probability distribution of the multioutput GP model can be written as
\begin{align*}
p(\{\bfy_d,\bff_d\}_{d=1}^D) = \prod_{d=1}^D p(\bfy_d|\bff_d)
p(\bff_1,\ldots, \bff_D).  
\end{align*}
The GP prior $p(\bff_1,\ldots, \bff_D)$
has a zero mean vector and a $(N D) \times (N D)$ covariance matrix
$\kMatrix_{\bff,\bff}$, where $\bff = (\bff_1,\ldots,\bff_D)$, which
consists of $N \times N$ blocks of the form
$\kMatrix_{\bff_d,\bff_{d'}}$. Elements of each block are given by $k_{f_d,f_{d'}}(\bfx,\bfx')$ 
for all possible values of $\bfx$. Each of such blocks is either a cross-covariance
or covariance matrix of pairs of outputs.

Prediction using the above GP model, as well as the maximization of
the marginal likelihood $p(\bfy) = N(\bfy|\bfzero,
\kMatrix_{\bff,\bff} + \bm{\Sigma})$, where $\bm{\Sigma} =
\text{diag}(\sigma_1^2 \mathbf{I}, \ldots, \sigma_D^2 \mathbf{I})$, requires $O(N^3
D^3)$ time and $O(N^2 D^2)$ storage which rapidly becomes infeasible
even when only few hundreds of outputs and data are considered.
Therefore approximate or sparse methods are needed in order to make
the above multioutput GP model practical.
  
\section{PITC-like approximation for Multiple Outputs Gaussian Processes}
\label{review:PITC}
Before we propose our variational sparse inference method for
multioutput GP regression in Section \ref{sec:sparsevarMultiOut}, we
review the sparse method proposed by
\citet{Alvarez:sparse2009}. This method is based on a likelihood
approximation. More precisely, each output function $y_{d} (\bfx)$ is
independent from the other output functions given the full-length of
each latent function $u_q(\bfx)$. This means, that the likelihood of
the data factorizes according to 
\begin{align*}
p(\bfy|u) = \prod_{d=1}^D p(\bfy_d|u) = \prod_{d=1}^D p(\bfy_d|\bff_d), 
\end{align*}
with $u=\{u_q\}_{q=1}^Q$ the set of latent functions. The sparse method in \citet{Alvarez:sparse2009}
makes use of this factorization by assuming that it remains valid even
when we are only allowed to exploit the information provided by a
finite set of function values, $\bfu_q$, instead of the full-length
function $u_q(\bfx)$ (which involves uncountably many points).
Let $\bfu_q$, for $q=1,\ldots,Q$, be a $K$-dimensional vector of
values from the function $u_q(\bfx)$ which are evaluated at the
inputs $\mathbf{Z}=\{\bfz_k\}_{k=1}^K$. These points are commonly referred to as
\emph{inducing inputs}. The vector $\bfu =
(\bfu_1,\ldots,\bfu_Q)$ denotes all these variables.  The sparse
method approximates the exact likelihood function
$p(\bfy|u)$ with the likelihood
\begin{align*}
p(\bfy|\bfu)&=\prod_{d=1}^Dp(\bfy_d|\bfu)=
\prod_{d=1}^D\mathcal{N}(\bfy_d|\bm{\mu}_{\mathbf{f}_d|\mathbf{u}},
\bm{\Sigma}_{\mathbf{f}_d|\mathbf{u}} + \sigma_d^2 I), 
\end{align*}
where
$\bm{\mu}_{\mathbf{f}_d|\mathbf{u}}=\kMatrix_{\mathbf{f}_d,\mathbf{u}}
\kMatrix^{-1}_{\mathbf{u},\mathbf{u}}\bfu$ and
$\bm{\Sigma}_{\mathbf{f}_d|\mathbf{u}}=\kMatrix_{\mathbf{f}_d,
  \mathbf{f}_d}-\kMatrix_{\mathbf{f}_d,
  \mathbf{u}}\kMatrix^{-1}_{\mathbf{u},\mathbf{u}}\kMatrix_{\mathbf{u},\mathbf{f}_d}$
are the mean and covariance matrices of the conditional GP priors
$p(\bff_d|\bfu)$. The matrix $\kMatrix_{\bfu,\bfu}$ is a block
diagonal covariance matrix where the $q$th block
$\kMatrix_{\bfu_q,\bfu_q}$ is obtained by evaluating $k_q
(\bfz,\bfz')$ at the inducing inputs $\mathbf{Z}$. Further, the
matrix $\kMatrix_{\bff_d,\bfu}$ has entries defined by the cross-covariance
function 
\begin{align*}
\text{Cov}[f_d(\bfx),u_q(\bfz)] = \int_{\mathcal{Z}} G_{d,q}(\bfx - \bfz')
k_{q}(\bfz',\bfz) d\bfz'. 
\end{align*}
The variables $\bfu$ follow the GP prior $p(\bfu) = N(\bfu|\bfzero, \kMatrix_{\bfu,\bfu})$ and can be
integrated out to give the following approximation to the exact
marginal likelihood:
\begin{equation}
p(\bfy|\bm{\theta}) =\mathcal{N}(\bfy|\mathbf{0},\mathbf{D} + \kMatrix_{\mathbf{f},\mathbf{u}}
\kMatrix^{-1}_{\mathbf{u},\mathbf{u}}\kMatrix_{\mathbf{u},\mathbf{f}}+\bm{\Sigma}).
\label{eq:PITCmarginalLik}
\end{equation}
Here, $\mathbf{D}$ is a block-diagonal matrix, where each block in the diagonal is given by $\kMatrix_{\mathbf{f}_d,
\mathbf{f}_d}-\kMatrix_{\mathbf{f}_d,\mathbf{u}}\kMatrix^{-1}_{\mathbf{u},\mathbf{u}}\kMatrix_{\mathbf{u},\mathbf{f}_d}$ for all $d$.
This approximate marginal likelihood represents exactly each diagonal 
(output-specific) block  
$\kMatrix_{\bff_d,\bff_d}$ while each off diagonal (cross-output) block
$\kMatrix_{\bff_d,\bff_{d'}}$ is approximated by the Nystr\"{o}m matrix 
$\kMatrix_{\mathbf{f}_d,\mathbf{u}}
\kMatrix^{-1}_{\mathbf{u},\mathbf{u}}\kMatrix_{\mathbf{u},\mathbf{f}_{d'}}$. 

The above sparse method has a similar structure to the PITC
approximation introduced for single-output regression
\citep{Quinonero:unifying05}. Because of this similarity,
\citet{Alvarez:sparse2009} call their multioutput sparse
approximation PITC as well.  Two of the properties of this PITC
approximation, which can be also its limitations, are:
\begin{enumerate}
\item It assumes that all latent functions in $u$ are
  smooth.
\item It is based on a modification of the initial full GP model. This
  implies that the inducing inputs $\mathbf{Z}$ are extra kernel
  hyparameters in the modified GP model.
\end{enumerate}
Because of point 1, the method is not applicable when the latent
functions are white noise processes. An important class of problems
where we have to deal with white noise processes arise in linear SDEs
where the above sparse method is currently not applicable there.
Because of 2, the maximization of the marginal likelihood in eq.\
(\ref{eq:PITCmarginalLik}) with respect to $(\mathbf{Z},\bftheta)$,
where $\bftheta$ are model hyperparameters, may be prone to
overfitting especially when the number of variables in $\mathbf{Z}$ is
large.  Moreover, fitting a modified sparse GP model implies that the
full GP model is not approximated in a systematic and rigorous way
since there is no distance or divergence between the two models
that is minimized

In the next section, we address point 1 above by introducing the
concept of variational inducing kernels that allow us to efficiently
sparsify multioutput GP models having white noise latent functions.
Further, these inducing kernels are incorporated into the variational
inference method of \citet{Titsias:variational09} (thus addressing
point 2) that treats the inducing inputs $\mathbf{Z}$ as well as other
quantities associated with the inducing kernels as variational
parameters. The whole variational approach provides us with a very
flexible, robust to overfitting, approximation framework that
overcomes the limitations of the PITC approximation.

\section{Sparse variational approximation
  \label{sec:sparsevarMultiOut}} 

In this section, we introduce the concept of variational inducing
kernels (VIKs). VIKs give us a way to define more general inducing
variables that have larger approximation capacity than the $\bfu$
inducing variables used earlier and importantly allow us to deal with
white noise latent functions.  To motivate the idea, we first
explain why the $\bfu$ variables can work when the latent functions
are smooth and fail when these functions become white noises.

In PITC, we assume each latent function $u_q(\bfx)$ is smooth and we
sparsify the GP model through introducing, $\bfu_q$, inducing
variables which are direct observations of the latent function,
$u_q(\bfx)$, at particular input points.  Because of the latent
function's smoothness, the $\bfu_q$ variables also carry information
about other points in the function through the imposed prior over the latent function. 
So, having observed $\bfu_q$ we can reduce the uncertainty of the whole function. 

\begin{figure*}[ht!]
  \begin{center}
    \subfigure[Latent function is smooth]{
      \includegraphics[width=0.42\textwidth]{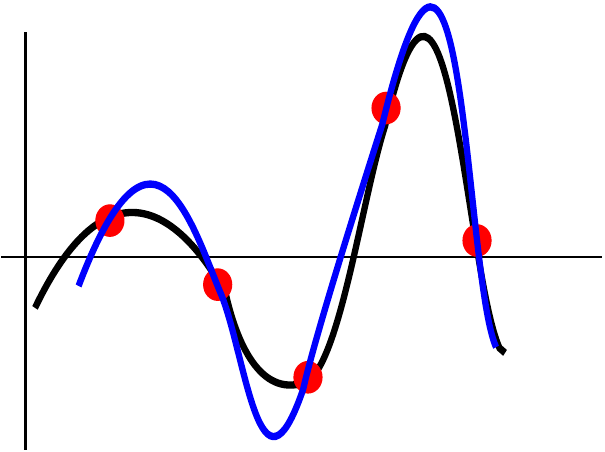}\label{fig:cartoonSmooth}}\hspace{2cm}
    \subfigure[Latent function is noise]{
      \includegraphics[width=0.42\textwidth]{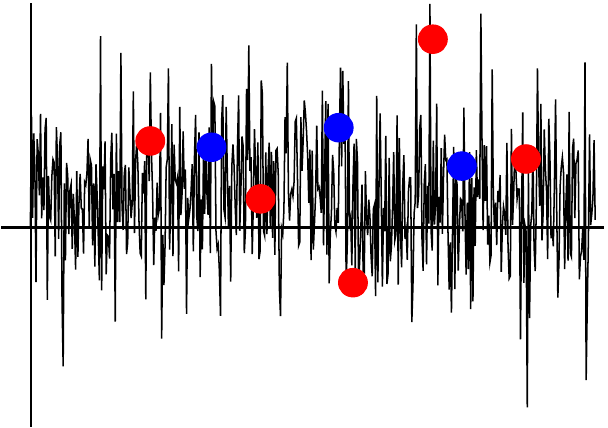}\label{fig:cartoonNoise}}
     \subfigure[Generation of an inducing function]{
       \includegraphics[width=0.98\textwidth]{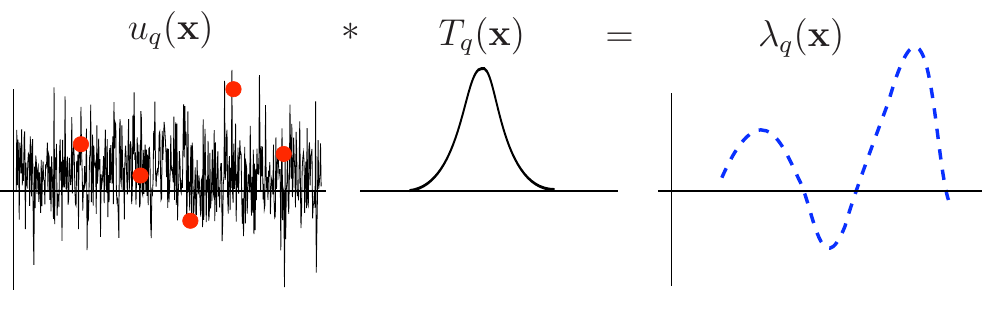} \label{fig:cartoonB}}
    {\caption{With a smooth latent function as in (a), we can use some inducing variables $\mathbf{u}_q$ (red dots) 
from the complete latent process $u_q(\mathbf{x})$ (in black) to generate smoothed versions  
(for example the one in blue), with uncertainty described by $p(u_q|\bfu_q)$.  
However, with a white noise latent function as in (b), choosing inducing variables $\mathbf{u}_q$ (red dots) from
the latent process (in black) does not give us a clue about other points (for example the blue dots).
In (c) the inducing function $\lambda_q(\mathbf{x})$ acts as a surrogate for a smooth function. Indirectly, 
it contains information about the inducing points and it can be used in the computation of the lower bound. In this 
context, the symbol $\ast$ refers to the convolution integral.
}
 \label{fig:cartoonA}}
  \end{center}
\end{figure*}

With the vector of inducing variables $\bfu$, if chosen to be
sufficiently large relative to the length scales of the latent
functions, we can efficiently represent the functions
$\{u_q(\bfx)\}_{q=1}^Q$ and subsequently variables $\bff$ which are
just convolved versions of the latent functions.\footnote{This idea is
  like a ``soft version'' of the Nyquist-Shannon sampling theorem. If
  the latent functions were bandlimited, we could compute exact results
  given a high enough number of inducing points. In general it won't be
  bandlimited, but for smooth functions low frecuency components will dominate over high frecuencies, which will 
quickly fade away.} When the reconstruction of $\bff$ from $\bfu$ is
perfect, the conditional prior $p(\bff|\bfu)$ becomes a delta function
and the sparse PITC approximation becomes exact. Figure
\ref{fig:cartoonSmooth} shows a cartoon description of a summarization of $u_q(\bfx)$ by $\bfu_q$.

In contrast, when some of the latent functions are {\em white noise}
processes the sparse approximation will fail. If $u_q(\bfz)$ is white
noise\footnote{Such a process can be thought as the ``time
  derivative'' of the Wiener process.} it has a covariance function
$\delta(\bfz - \bfz')$. Such processes naturally arise in the
application of {\em stochastic differential equations} (see section
\ref{Sec:SDE}) and are the ultimate non-smooth processes where two
values $u_q(\bfz)$ and $u_q(\bfz')$ are uncorrelated when $\bfz \neq
\bfz'$. When we apply the sparse approximation a vector of
``white-noise'' inducing variables $\bfu_q$ does not carry information
about $u_q(\bfz)$ at any input $\bfz$ that differs from all inducing
inputs $\mathbf{Z}$. In other words there is no additional information
in the conditional prior $p(u_q(\bfz)|\bfu_q)$ over the unconditional
prior $p(u_q(\bfz))$. Figure \ref{fig:cartoonNoise} shows a pictorial
representation. The lack of structure makes it impossible to exploit
the correlations in the standard sparse methods like
PITC.\footnote{Returning to our sampling theorem analogy, the white
  noise process has infinite bandwidth. It is therefore impossible to
  represent it by observations at a few fixed inducing points.}

Our solution to this problem is the following. We will define a more
powerful form of inducing variable, one based not around the latent
function at a point, but one given by the convolution of the latent
function with a smoothing kernel. More precisely, let us replace each
inducing vector $\bfu_q$ with the variables $\bflambda_q$ which are
evaluated at the inputs $\mathbf{Z}$ and are defined according to
\begin{equation}
\lambda_q(\bfz) =  \int T_q (\bfz - \mathbf{v}) u_q (\mathbf{v}) \dif \mathbf{v}, 
\label{eq:VIK}
\end{equation}
where $T_q(\bfx)$ is a smoothing kernel (e.g. Gaussian) which we call
the \emph{inducing kernel} (IK).  This kernel is not necessarily
related to the model's smoothing kernels.  These newly defined
inducing variables can carry information about $u_q(\bfz)$ not only at
a single input location but from the entire input space.
Figure \ref{fig:cartoonB} shows how the inducing kernel 
generates the artificial construction $\lambda_q(\mathbf{x})$, that shares some ligth over the, otherwise,
obscure inducing points. 
We can even allow a separate IK for each inducing point, this is, if
the set of inducing points is $\mathbf{Z}=\{\bfz_k\}_{k=1}^K$, then
\begin{align*}
\lambda_q(\bfz_k) = \int T_{q,k} (\mathbf{z}_k - \mathbf{v}) u_q
(\mathbf{v}) \dif \mathbf{v}, 
\end{align*}
with the advantage of associating to each
inducing point $\bfz_k$ its own set of adaptive parameters in
$T_{q,k}$. For the PITC approximation, this adds more hyperparameters
to the likelihood, perhaps leading to overfitting. However, in the
variational approximation we define all these new parameters as
variational parameters and therefore they do not cause the model to
overfit. We use the notation $\lambda$ to refer to the set of inducing functions $\{\lambda_q\}_{q=1}^Q$.

If $u_q(\bfz)$ has a white noise 
\footnote{It is straightforward to
  generalize the method for rough latent functions that are not white
  noise or to combine smooth latent functions with white noise.} 
GP prior the covariance function for $\lambda_q(\bfx)$ is
\begin{equation}
  \text{Cov}[\lambda_q(\bfx),\lambda_q(\bfx')] 
  = \int T_{q} (\bfx - \bfz) T_q(\bfx' - \bfz)  
  \mathrm{d} \bfz
\label{eq:covlambda}
\end{equation}
and the cross-covariance function between $f_d(\bfx)$ and
$\lambda_q(\bfx')$ is
\begin{equation}
  \text{Cov}[f_d(\bfx),\lambda_q(\bfx')] = 
  \int G_{d,q} (\bfx - \bfz) T_q(\bfx' - \bfz)  d\bfz.
\label{eq:crossCovlambdaf}
\end{equation} 
Notice that this cross-covariance function, unlike the case of $\bfu$
inducing variables, maintains a weighted integration over the whole input
space.  This implies that a single inducing variable $\lambda_q(\bfx)$
can properly propagate information from the full-length process
$u_q(\bfx)$ into the set of outputs $\bff$.

It is possible to combine the IKs defined above with the PITC
approximation of \citet{Alvarez:sparse2009}, but in this paper our
focus will be on applying them within the variational framework of
\citet{Titsias:variational09}. We therefore refer to the kernels as
variational inducing kernels (VIKs).
  
\section{Variational inference for sparse multiple output Gaussian Processes. } 

We now extend the variational inference method of
\citet{Titsias:variational09} to deal with multiple outputs and
incorporate them into the VIK framework.

We compactly write the joint probability model
$p(\{\bfy_d,\bff_d\}_{d=1}^D)$ as $p(\bfy,\bff) = p(\bfy | \bff)
p(\bff)$. The first step of the variational method is to augment this
model with inducing variables. For our purpose, suitable inducing
variables are defined through VIKs.  More precisely, let $\bflambda =
(\bflambda_1, \ldots, \bflambda_Q)$ be the whole vector of inducing
variables where each $\bflambda_q$ is a $K$-dimensional vector of
values obtained according to eq.\ (\ref{eq:VIK}). The role of $\bflambda_q$
is to carry information about the latent function $u_q(\bfz)$. Each
$\bflambda_q$ is evaluated at the inputs $\mathbf{Z}$ and has its own
VIK, $T_q(\bfx)$, that depends on parameters $\bftheta_{T_q}$. We denote these parameters as $\bm{\Theta}=
\{\bm{\theta}_{T_q}\}_{q=1}^Q$.

The $\bflambda$ variables augment the GP model according to
\begin{align*}
p(\bfy,\bff,\bflambda) = p(\bfy|\bff) p(\bff|\bflambda)
p(\bflambda).  
\end{align*}
Here, $p(\bflambda) = \mathcal{N}(\bflambda|\bfzero,
\kMatrix_{\bflambda, \bflambda})$ and $\kMatrix_{\bflambda,
  \bflambda}$ is a block diagonal matrix where each block
$\kMatrix_{\bflambda_q,\bflambda_q}$ in the diagonal is obtained by evaluating the
covariance function in eq.\ (\ref{eq:covlambda}) at the inputs
$\mathbf{Z}$. Additionally, $p(\bff|\bflambda) = \mathcal{N}(\bff|
\kMatrix_{\bff,\bflambda} \kMatrix_{\bflambda,\bflambda}^{-1}
\bflambda, \kMatrix_{\bff,\bff} - \kMatrix_{\bff,\bflambda}
\kMatrix_{\bflambda, \bflambda}^{-1} \kMatrix_{\bflambda, \bff})$
where the cross-covariance $\kMatrix_{\bff,\bflambda}$ is computed
through eq.\ (\ref{eq:crossCovlambdaf}). Because of the consistency
condition $\int p(\bff|\bflambda) p(\bflambda) \dif\bflambda =
p(\bff)$, performing exact inference in the above augmented model is
equivalent to performing exact inference in the initial GP
model. Crucially, this holds for any values of the {\em augmentation}
parameters $(\mathbf{Z},\bm{\Theta})$. This is the key
property that allows us to turn these augmentation parameters into
variational parameters by applying approximate sparse inference.
 
Our method now follows exactly the lines of
\citet{Titsias:variational09} (in appendix \ref{appendix:bound} we present a detailed derivation of the bound based
on the set of latent functions $u_q(\mathbf{x})$).
We introduce the variational
distribution $q(\bff,\bflambda) = p(\bff| \bflambda) \phi(\bflambda)$,
where $p(\bff| \bflambda)$ is the conditional GP prior defined earlier
and $\phi(\bflambda)$ is an arbitrary variational distribution.  By
minimizing the KL divergence between $q(\bff,\bflambda)$ and the true
posterior $p(\bff,\bflambda|\bfy)$, we can compute the following
Jensen's lower bound on the true log marginal likelihood:
\[
\mathcal{F}_{V}(\mathbf{Z},\bm{\Theta})=\log\mathcal{N}\left(\bfy|\mathbf{0},\kMatrix_{\bff,\bflambda}\kMatrix_{\bflambda,\bflambda}^{-1}\kMatrix_{\bflambda,\bff}+\bm{\Sigma}
\right)-\frac{1}{2}\tr\left(\bm{\Sigma}^{-1}\widetilde{\kMatrix}\right),
\]
where $\bm{\Sigma}$ is the covariance function associated with the
additive noise process and $\widetilde{\kMatrix} =
\kMatrix_{\bff,\bff}-\kMatrix_{\bff,\bflambda}\kMatrix_{\bflambda,\bflambda}^{-1}
\kMatrix_{\bflambda,\bff}$.  Note that this bound consists of two
parts.  The first part is the log of a GP prior with the only
difference that now the covariance matrix has a particular low rank
form. This form allows the inversion of the covariance matrix to take
place in $O(NDK^2)$ time rather than $O(N^3 D^3)$. The second part can be seen as a penalization 
term that regulates the estimation of the parameters. Notice also that
only the diagonal of the exact covariance matrix
$\kMatrix_{\bff,\bff}$ needs to be computed. Overall, the computation
of the bound can be done efficiently in $O(NDK^2)$ time.

The bound can be maximized with respect to all parameters of the
covariance function; both model parameters and variational
parameters. The variational parameters are the inducing inputs
$\mathbf{Z}$ and the parameters $\bftheta_{T_q}$ of each VIK which are
rigorously selected so that the KL divergence is minimized. In fact
each VIK is also a variational quantity and one could try different
forms of VIKs in order to choose the one that gives the best lower
bound.

The form of the bound is very similar to the projected process
approximation, also known as Deterministic Training Conditional approximation (DTC)
\citep{Csato:sparse00,Seeger:fast03,Rasmussen:book06}.  However, the
bound has an additional trace term that penalizes the movement of
inducing inputs away from the data. This term converts the DTC
approximation to a lower bound and prevents overfitting. In what
follows, we refer to this approximation as DTCVAR, where the VAR
suffix refers to the variational framework.

The predictive distribution of a vector of test points, 
$\mathbf{y}_*$ given the training data can also be found to be
\begin{align*}
p\left(\bfy_{*}|\bfy,\mathbf{X},\mathbf{Z}\right) =
\mathcal{N}\left(\bfy_{*}|\bm{\mu}_{\bfy_{*}},\bm{\Sigma}_{\bfy_{*}}\right),
\end{align*}
with
$\bm{\mu}_{\bfy_{*}}=\kMatrix_{\bff_{*}\bflambda}\mathbf{A}^{-1}\kMatrix_{\bflambda\bff}
\bm{\Sigma}^{-1}\bfy$
and
$\bm{\Sigma}_{\bfy_{*}}=\kMatrix_{\bff_{*}\bff_{*}}-\kMatrix_{\bff_{*}\bflambda}
\left(\kMatrix_{\bflambda\bflambda}^{-1}-\mathbf{A}^{-1}\right)\kMatrix_{\bflambda\bff_{*}}+\bm{\Sigma_*}$
and
$\mathbf{A}=\kMatrix_{\bflambda,\bflambda}+\kMatrix_{\bflambda,\bff}\bm{\Sigma}^{-1}\kMatrix_{\bff,\bflambda}$. Predictive
means can be computed in $O(NDK)$ whereas predictive variances require
$O(NDK^2)$ computation. 

\section{Experiments}
\label{Sec:Simulations}

We present results of applying the method proposed for two real-world
datasets that will be described in short.  We compare the results
obtained using PITC, the intrinsic coregionalization model
(ICM)\footnote{The intrinsic coregionalization model is a particular
  case of the linear model of coregionalization with one latent
  function \citep{Goovaerts:book97}. See equation \eqref{eq:lmc:model} with $Q=1$.}  employed in
\citep{Bonilla:multi07} and the method using variational inducing
kernels. For PITC we estimate the parameters through the maximization
of the approximated marginal likelihood of equation
\eqref{eq:PITCmarginalLik} using a scaled-conjugate gradient
method. We use one latent function and both the covariance function of
the latent process, $k_q(\bfx,\bfx')$, and the kernel smoothing
function, $G_{d,q}(\bfx)$, follow a Gaussian form, this is
\begin{align*}
k(\mathbf{x}, \mathbf{x}')=\mathcal{N}(\mathbf{x} - \mathbf{x}'|\mathbf{0}, \mathbf{C}), 
\end{align*}
where $\mathbf{C}$ is a diagonal
matrix.  For the DTCVAR approximations, we maximize the variational
bound $\mathcal{F}_V$. Optimization is also performed using scaled conjugate
gradient. We use one white noise latent function and a corresponding
inducing function. The inducing kernels and the model kernels follow
the same Gaussian form as in the PITC case.  Using this form for the
covariance or kernel, all convolution integrals are solved
analytically.

\subsection{Exam score prediction}

In this experiment the goal is to predict the exam score obtained by a
particular student belonging to a particular school. The data comes
from the Inner London Education Authority (ILEA).\footnote{Data
  is available at
  \url{http://www.cmm.bristol.ac.uk/learning-training/multilevel-m-support/datasets.shtml}} 
It consists of examination records from 139 secondary schools in years
1985, 1986 and 1987. It is a random $50\%$ sample with 15362
students. The input space consists of features related to each
student and features related to each school. From the multiple
output point of view, each school represents one output and the exam
score of each student a particular instantiation of that output.

We follow the same preprocessing steps employed in
\citep{Bonilla:multi07}. The only features used are the
student-dependent ones (year in which each student took the exam, gender, VR band and
ethnic group), which are categorial variables. Each of them
is transformed to a binary representation. For example, the possible
values that the variable year of the exam can take are 1985, 1986 or
1987 and are represented as $100$, $010$ or $001$. The transformation
is also applied to the variables gender (two binary variables), VR
band (four binary variables) and ethnic group (eleven binary
variables), ending up with an input space with dimension $20$. The
categorial nature of data restricts the input space to $202$ unique
input feature vectors. However, two students represented by the same
input vector $\mathbf{x}$ and belonging both to the same school $d$,
can obtain different exam scores. To reduce this noise in the data, we
follow \citet{Bonilla:multi07} in taking the mean of the observations that, 
within a school, share the same input vector and use
a simple heteroskedastic noise model in which the variance for each of
these means is divided by the number of observations used to compute
it. 
The performance measure employed is the percentage of explained variance
defined as the total variance of the data minus the sum-squared error
on the test set as a percentage of the total data variance. It can be
seen as the percentage version of the coefficient of determination
between the test targets and the predictions. The performance measure is computed for ten repetitions 
with $75\%$ of the data in the training set and $25\%$ of the data in the test set.

\begin{figure}[ht!]
  \begin{center}
  \includegraphics[width=0.98\textwidth]{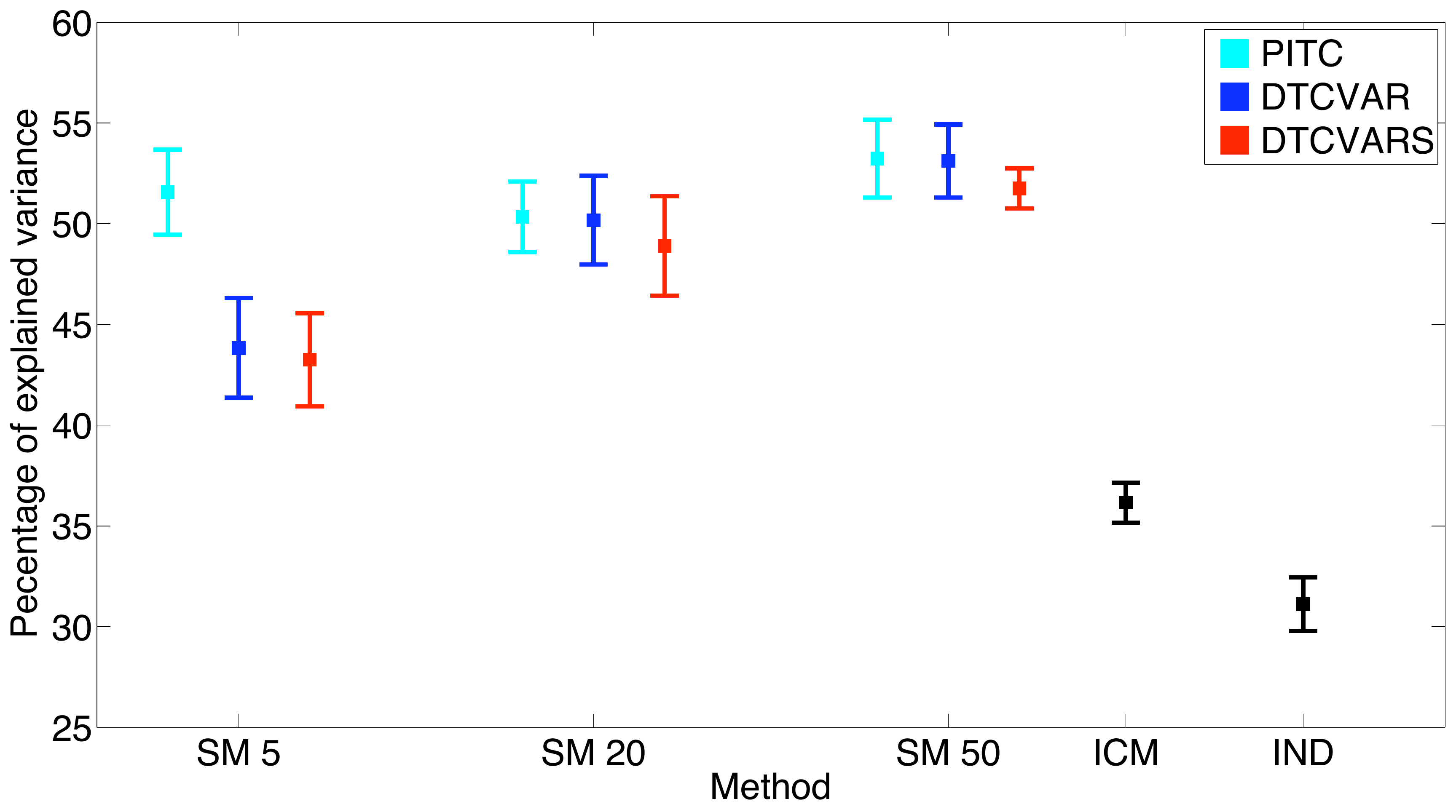}
  \small{\caption{Exam score prediction results for the school dataset. Results include the mean of the percentage of
explained variance of ten repetitions of the experiment, together with one standard deviation. 
In the bottom, SM X stands for 
sparse method with X inducing points, DTCVAR refers to the DTC variational
approximation with one smoothing kernel and DTCVARS to the same approximation using as many inducing kernels as 
inducing points. Results 
using the ICM model and independent GPs (appearing as IND in the figure) have also been included. 
}}\label{fig:results:school}
 \end{center}
\end{figure}

Figure \ref{fig:results:school} shows results using PITC, DTCVAR with
one smoothing kernel and DTCVAR with as many inducing kernels as
inducing points (DTCVARS in the figure). For $50$ inducing points all
three alternatives lead to approximately the same results. PITC keeps
a relatively constant performance for all values of inducing points, while the
DTCVAR approximations increase their performance as the number of
inducing points increase. This is consistent with the expected 
behaviour of the DTCVAR methods, since the trace term penalizes the
model for a reduced number of inducing points.  Notice that all the
approximations outperform independent GPs and the best result of the
intrinsic coregionalization model presented in
\citep{Bonilla:multi07}.

\subsection{Compiler prediction performance.} 

In this dataset the outputs correspond to the speed-up of 11 C
programs after some transformation sequence has been applied to
them. The speed-up is defined as the execution time of the original
program divided by the execution time of the transformed program. The
input space consists of 13-dimensional binary feature vectors, where
the presence of a one in these vectors indicates that the program has
received that particular transformation.  The dataset contains 88214
observations for each output and the same number of input vectors. All
the outputs share the same input space. Due to technical requirements,
it is important that the prediction of the speed-up for the particular
program is made using few observations in the training set. We compare
our results to the ones presented in \citep{Bonilla:multi07} and use
$N=16$, $32$, $64$ and $128$ for the training set. The remaining
$88214-N$ observations are used for testing, employing as performance
measure the mean absolute error. The experiment is repeated ten times
and standard deviations are also reported. We only include results for
the average performance over the 11 outputs.

\begin{figure}[ht!]
  \begin{center}
     {
      \includegraphics[width=0.98\textwidth]{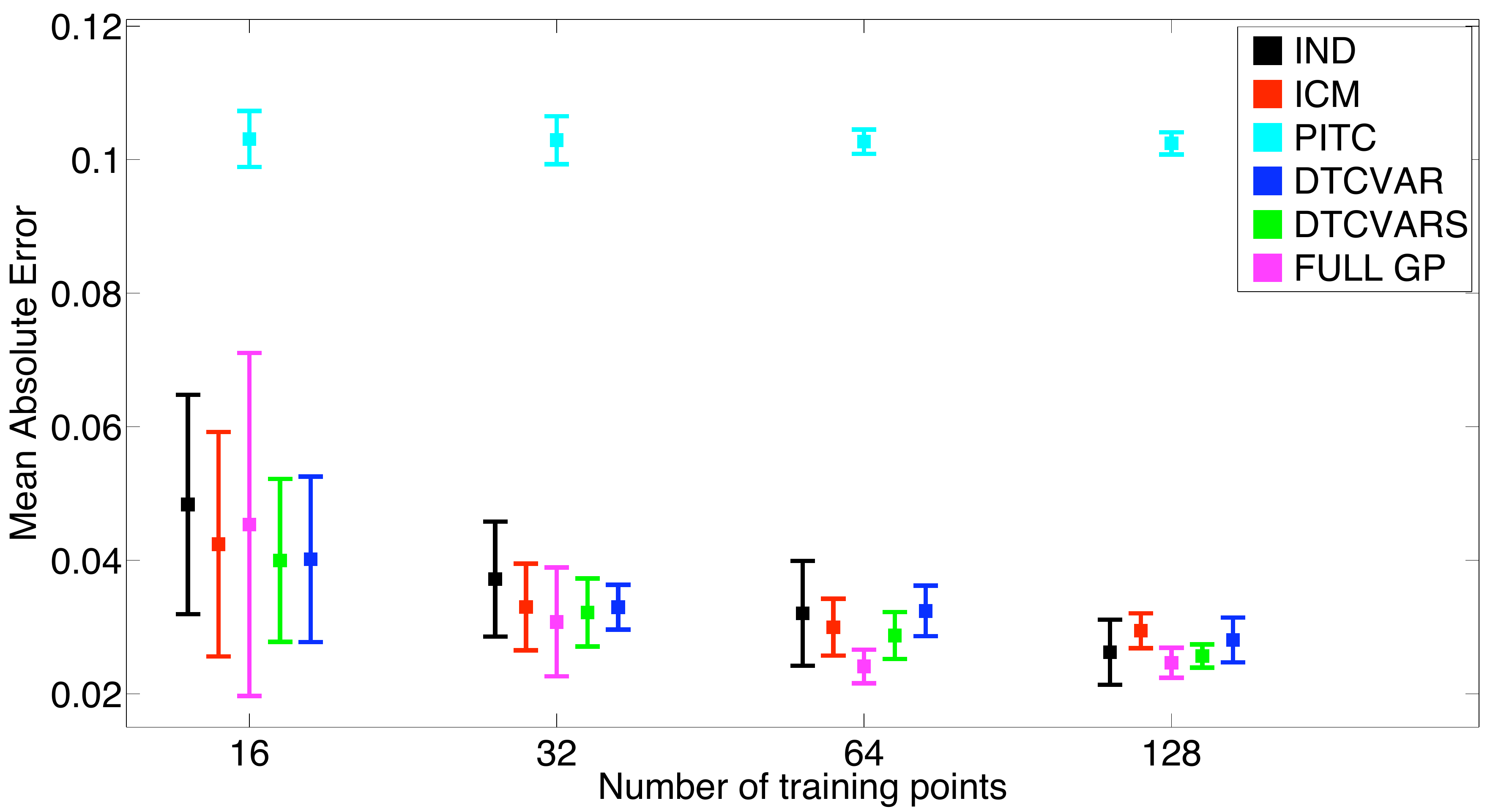}}
    {\caption{Mean absolute error and standard deviation over
        ten repetitions of the compiler experiment as a function of
        the training points. IND stands for independent GPs, ICM
        stands for intrinsic coregionalization model, DTCVAR refers to
        the DTCVAR approximation using one inducing kernel, DTCVARS
        refers to the DTCVAR approximation using as many inducing
        kernels as inducing points and FULL GP stands for the GP for
        the multiple outputs without any approximation.  
        } \label{fig:resultsCompiler}}
  \end{center}
\end{figure}

Figure \ref{fig:resultsCompiler} shows the results of applying
independent GPs (IND in the figure), the intrinsic coregionalization
model (ICM in the figure), PITC, DTCVAR with one inducing kernel
(DTCVAR in the figure) and DTCVAR with as many inducing kernels as
inducing points (DTCVARS in the figure). Since the training sets are
small enough, we also include results of applying the GP generated
using the full covariance matrix of the convolution construction (see
FULL GP in the figure). We repeated the experiment for different values of $K$, but show results only for $K=N/2$.
Results for ICM and IND were obtained from
\citep{Bonilla:multi07}.
In general, the DTCVAR variants outperform the ICM method, and the
independent GPs for $N=16,\;32$ and $64$.  In this case, using as many
inducing kernels as inducing points improves in average the
performance. All methods, including the independent GPs are better
than PITC. The size of the test set encourages the application of
the sparse methods: for $N=128$, making the prediction of the whole
dataset using the full GP takes in average $22$ minutes while the
prediction with DTCVAR takes $0.65$ minutes. 
Using more inducing kernels improves the performance, but also makes
the evaluation of the test set more expensive. For DTCVARS, the
evaluation takes in average $6.8$ minutes.
Time results are average results over the ten repetitions.

\section{Stochastic Latent Force Models for Financial Data}
\label{Sec:SDE}

The starting point of stochastic differential equations is a
stochastic version of the equation of motion, which is called Langevin's
equation:
\begin{equation}
\frac{\mathrm{d} f(t)}{\mathrm{d} t} =  - C f(t)  +  S u(t), 
\label{eq:Langevin}
\end{equation}
where $f(t)$ is the velocity of the particle, 
$ - Cf(t)$  is a systematic friction term, $u(t)$ 
is a random fluctuation external force, i.e.\  
white noise, and $S$ indicates the sensitivity of the ouput to the random fluctuations. 
In the mathematical probability literature, 
the above is written more rigorously as  
$ \mathrm{d} f(t) = - C f(t) \mathrm{d} t + S \mathrm{d} W (t)$ 
where $W (t)$ is the Wiener process (standard Brownian motion).
Since $u(t)$ is a Gaussian process and the equation is linear,
 $f(t)$ must be also a Gaussian process which 
turns out to be the Ornstein-Uhlenbeck (OU) process.  

Here, we are interested in extending the Langevin equation to model
multivariate time series. The way that the model in 
(\ref{eq:Langevin}) is extended is by adding more output signals and
more external forces. The forces can be either smooth (systematic or
drift-type) forces or white noise forces.  Thus, we obtain
\begin{align}\label{eq:basic:SDE}
\frac{\mathrm{d} f_d (t)}{\mathrm{d} t} &=  - D_d f_d (t)  + \sum_{q=1}^Q S_{d,q} u_q (t),
\end{align}
where $f_d (t)$ is the $d$th output signal. Each $u_q(t)$ can be either a smooth latent force that is
assigned a GP prior with covariance function $k_q (t,t')$ or a white
noise force that has a GP prior with covariance function $\delta(t -
t')$. That is, we have a composition of $Q$ latent forces, where $Q_s$
of them correspond to smooth latent forces and $Q_o$ correspond to
white noise processes. The intuition behind this combination of input
forces is that the smooth part can be used to represent medium/long
term trends that cause a departure from the mean of the output
processes, whereas the stochastic part is related to short term
fluctuations around the mean.  A model that employs $Q_s=1$ and
$Q_o=0$ was proposed by \citet{Lawrence:gpsim2007a} to
describe protein transcription regulation in a single input motif
(SIM) gene network.

Solving the differential equation \eqref{eq:basic:SDE}, we obtain
\begin{align*}
f_d(t) = e^{-D_d t} f_{d,0} + \sum_{q=1}^Q S_{d,q} \int_{0}^t e^{ -D_d (t-z) } u_q (z) dz,
\end{align*} 
where $f_{d,0}$ arises from the initial
condition. This model now is a special case of the multioutput
regression model discussed in sections 1 and 2 where each output
signal $y_d(t) = f_d(t) + \epsilon$ has a mean function $e^{-D_d t}
f_{d,0}$ and each model kernel $G_{d,q}(\bfx)$ is equal to $S_{d,q}
e^{ - D_d (t-z) }$. The above model can be also viewed as a stochastic
latent force model (SLFM) following the work of \citet{Alvarez:lfm09}.

\subsection*{Latent market forces}

\begin{figure*}[ht!]
  \centering \subfigure[CAD: Real data and prediction]{
    \resizebox{0.48\textwidth}{!}{\includegraphics{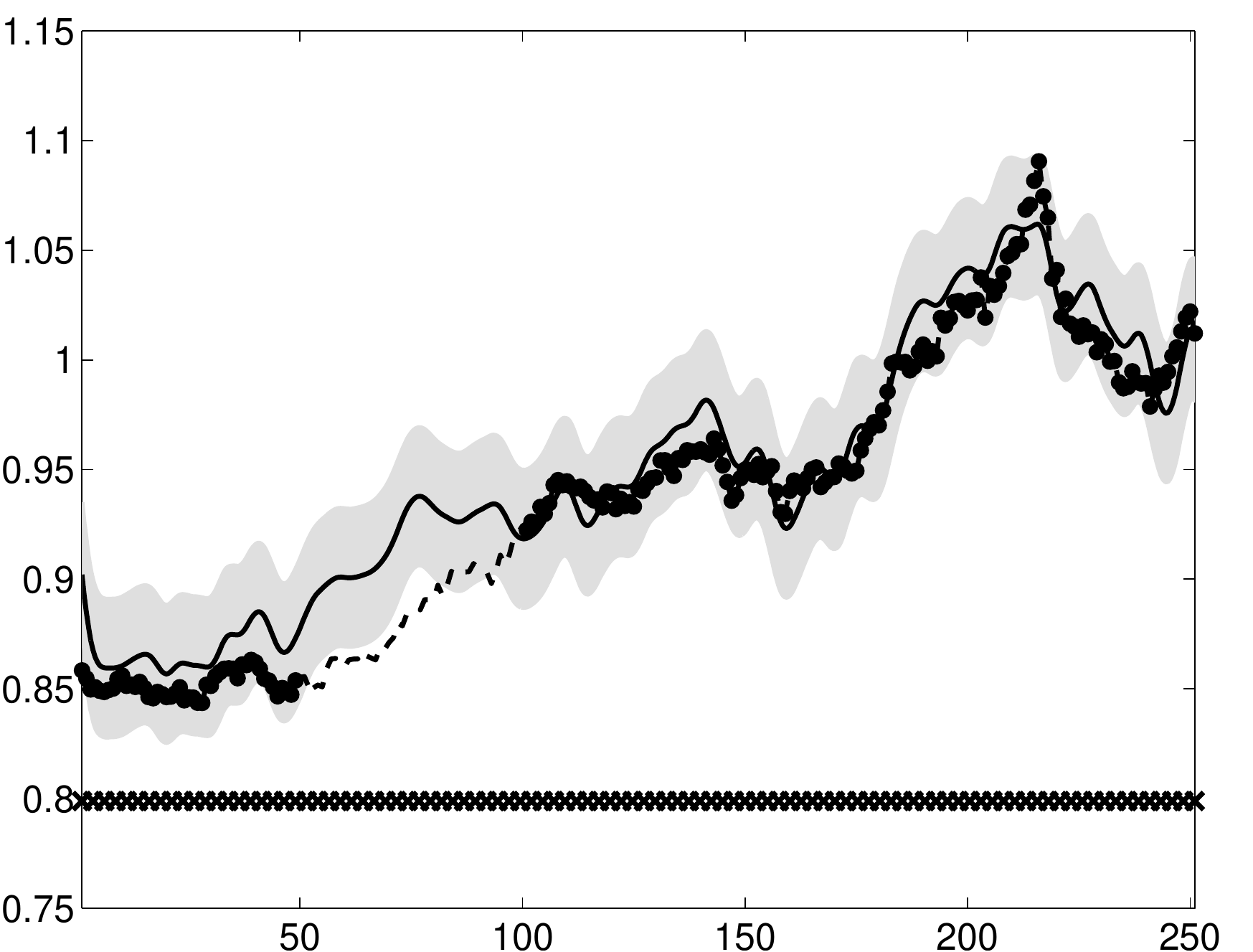}}}
  \subfigure[JPY: Real data and prediction]{
    \resizebox{0.48\textwidth}{!}{\includegraphics{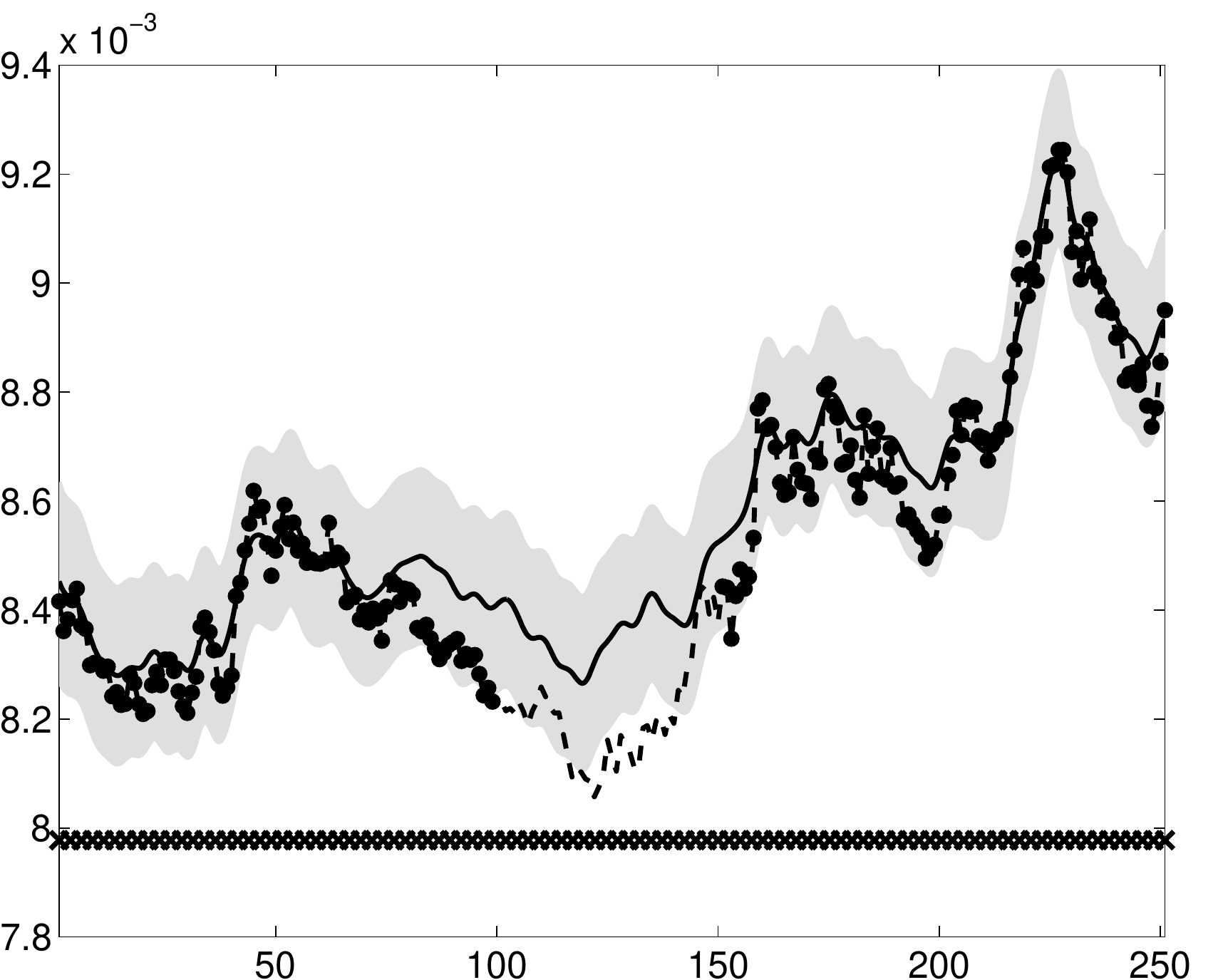}}}
  \subfigure[AUD: Real data and prediction]{
    \resizebox{0.48\textwidth}{!}{\includegraphics{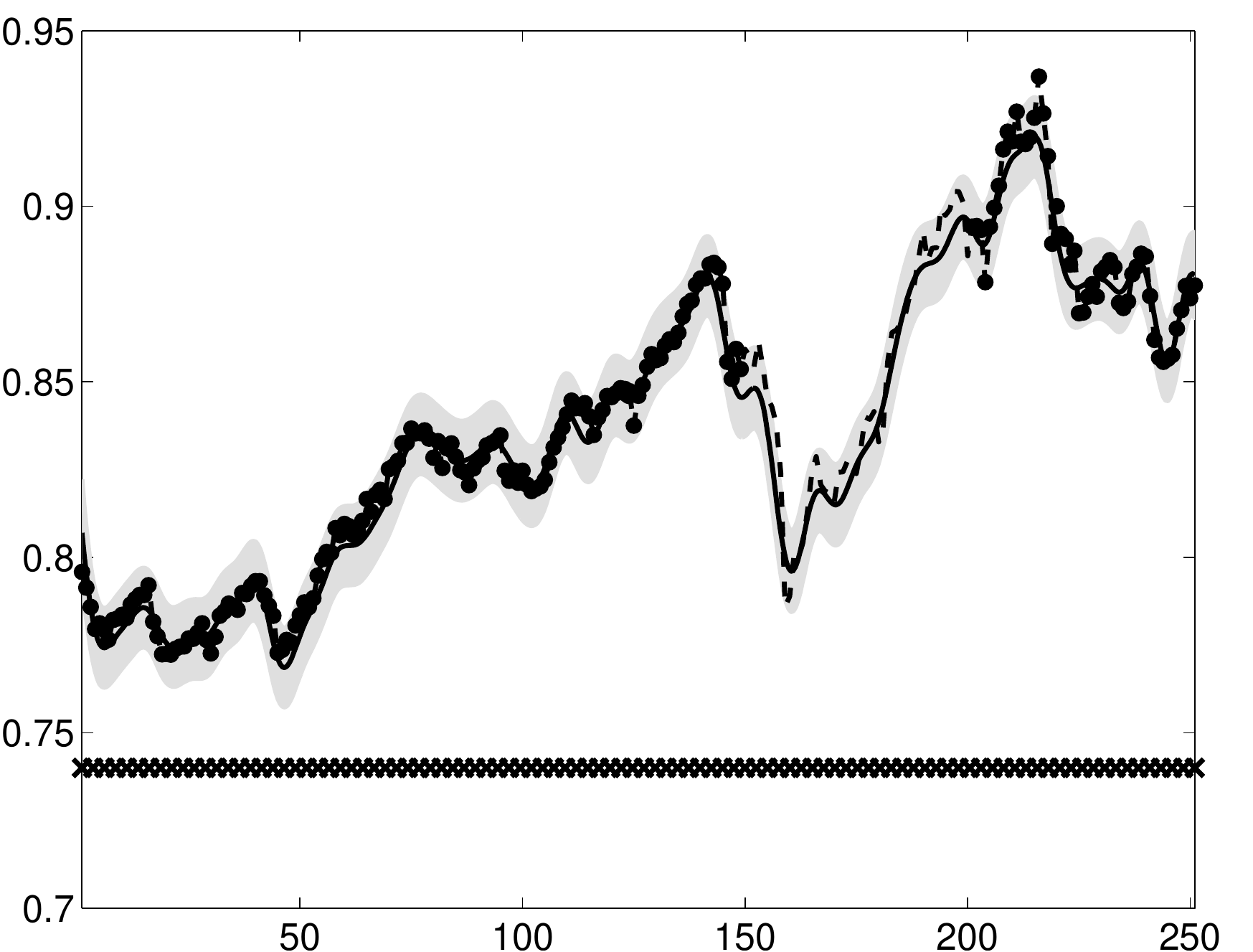}}}
      \caption{Predictions from the model with $Q_s=1$ and $Q_o=3$ are shown as solid 
     lines for the mean and grey bars for error bars at 2 standard
      deviations. 
     For CAD, JPY and AUD the data was artificially held out. 
     The true values are shown as a dotted line. 
      Crosses on the $x$-axes of all plots show the locations of the inducing inputs.}
  \label{fig:FXDataResults}
\end{figure*}

The application considered is the inference of missing data in a
multivariate financial data set: the foreign exchange rate w.r.t. the
dollar of 10 of the top international currencies (Canadian Dollar
[CAD], Euro [EUR], Japanese Yen [JPY], Great British Pound [GBP],
Swiss Franc [CHF], Australian Dollar [AUD], Hong Kong Dollar [HKD],
New Zealand Dollar [NZD], South Korean Won [KRW] and Mexican Peso
[MXN]) and 3 precious metals (gold [XAU], silver [XAG] and platinum
[XPT]).\footnote{Data is available at
  \url{http://fx.sauder.ubc.ca/data.html}).} We considered all the
data available for the calendar year of 2007 (251 working days). In
this data there are several missing values: XAU, XAG and XPT have 9, 8
and 42 days of missing values respectively. On top of this, we also
introduced artificially long sequences of missing data. Our objective
is to model the data and test the effectiveness of the model by
imputing these missing points. We removed a test set from the data by
extracting contiguous sections from 3 currencies associated with very
different geographic locations: we took days 50--100 from CAD, days
100--150 from JPY and days 150--200 from AUD. The remainder of the
data comprised the training set, which consisted of 3051 points, with
the test data containing 153 points. For preprocessing we removed the
mean from each output and scaled them so that they all had unit
variance.

It seems reasonable to suggest that the fluctuations of the 13
correlated financial time-series are driven by a smaller number of
latent market forces. We therefore modelled the data with up to six
latent forces which could be noise or smooth GPs. The GP priors for
the smooth latent forces are assumed to have a squared exponential
covariance function, 
\begin{align*}
k_{q}(t, t^\prime) = \frac{1}{\sqrt{2\pi\ell_q^2}} \exp\bigg(-\frac{(t-t^\prime)^2}{2\ell_q^2}\bigg),
\end{align*}
where the hyperparameter $\ell_q$ is known as the lengthscale.

We present an example with $Q=4$. For this value of $Q$, we consider all the possible combinations of
$Q_o$ and $Q_s$.  The training was performed in all cases by
maximizing the variational bound using the scale conjugate gradient
algorithm until convergence was achieved.
The best performance in terms of achiving the highest value for $\mathcal{F}_V$ was obtained for  
$Q_s=1$ and $Q_o=3$. We compared against the 
LMC model for different values of the latent 
functions in that framework. While our best model resulted in an standardized mean square error of $0.2795$, the best 
LMC (with $Q$=2) resulted in $0.3927$.  
We plotted predictions from the latent market force model to characterize the performance when 
filling in missing 
data. In figure \ref{fig:FXDataResults} we show the output signals obtained using the model with the highest 
bound ($Q_s=1$ and $Q_o=3$) for CAD, JPY and AUD. Note that the model performs better at capturing the deep drop in AUD 
than it does at capturing fluctuations in CAD and JPY. 

\section{Conclusions}

We have presented a variational approach to sparse approximations in
convolution processes. Our main focus was to provide efficient
mechanisms for learning in multiple output Gaussian processes when the
latent function is fluctuating rapidly. In order to do so, we have introduced the concept
of inducing function, which generalizes the idea of inducing point, traditionally employed in sparse GP methods.
The approach extends the
variational approximation of \citet{Titsias:variational09} to the
multiple output case. Using our approach we can perform efficient
inference on latent force models which are based around stochastic
differential equations, but also contain a smooth driving force. Our
approximation is flexible enough and has been shown to be applicable to a wide
range of data sets, including high-dimensional ones. 

\subsection*{Acknowledgements}

The authors would like to thank Edwin Bonilla for his valuable
feedback with respect to the exam score prediction example and the compiler dataset example. We also thank the authors
of \citet{Bonilla:multi07} who kindly made the compiler dataset available. DL has 
been partly financed by Comunidad de Madrid (project PRO-MULTIDIS-CM,
S-0505/TIC/0233), and by the Spanish government (CICYT project
TEC2006-13514-C02-01 and researh grant JC2008-00219).
MA and NL have been financed by a Google Research Award ``Mechanistically Inspired Convolution
Processes for Learning'' and MA, NL and MT have been financed by
EPSRC Grant No EP/F005687/1 ``Gaussian Processes for Systems Identification with Applications in Systems Biology''.

\appendix

\section{Variational Inducing Kernels}\label{appendix:bound}
Recently, a method for variational sparse approximation
for Gaussian processes learning was introduced in \citet{Titsias:variational09}.
In this appendix, we apply this methodology to a multiple output Gaussian
process where the outputs have been generated through a so called
convolution process. For learning the parameters of the kernels involved,
a lower bound for the true marginal can be maximized. This lower bound
has similar form to the marginal likelihood of the Deterministic Training
Conditional (DTC) approximation plus an extra term which involves
a trace operation. The computational complexity grows as $\mathcal{O}(NDK^{2})$
where $N$ is the number of data points per output, $D$ is the number
of outputs and $K$ the number of inducing variables. 

\subsection{Computation of the lower bound}\label{appendix:computation:bound}

Given target data $\bfy$ and inputs $\mathbf{X}$, the marginal likelihood
of the original model is given by integrating over the latent function\footnote{Strictly speaking, the  
distributions associated to $u$ correspond to random
signed measures, in particular, Gaussian measures.}
\[
p(\bfy|\mathbf{X})=\int_{u}p(\bfy|u,\mathbf{X})p(u)\dif{u}.\]
The prior over $u$ is expressed as 
\[
p(u)=\int_{\bflambda}p(u|\bflambda)p(\bflambda)\dif\bflambda.
\]
The augmented joint model can then be expressed as 
\begin{align*}
p\left(\mathbf{y},u,\bflambda\right)&=p(\mathbf{y}|u)p(u|\bflambda)p(\bflambda).
\end{align*}
With the inducing function $\bflambda$, the marginal likelihood takes the form
\[
p(\bfy|\mathbf{X})=\int_{u,\bflambda}p(\bfy|u,\mathbf{X})p(u|\bflambda)p(\bflambda)\dif{\bflambda}\,
\dif{u}.
\]
 Using Jensen's inequality, we use the following variational bound
on the log likelihood, 
\[
\mathcal{F}_{V}(\mathbf{Z},\bm{\Theta},\phi(\bflambda))=\int_{u,\bflambda}q(u,\bflambda)\log\frac{p(\bfy|u,\mathbf{X})
p(u|\bflambda)p(\bflambda)}{q(u,\bflambda)}\dif{\bflambda}\,\dif{u},
\]
 where we have introduced $q(u,\bflambda)$ as the variational
approximation to the posterior. Following \citet{Titsias:variational09}
we now specify that the variational approximation should be of the
form \[
q(u,\bflambda)=p(u|\bflambda)\phi(\bflambda).\]
 We can write our bound as%
\begin{align*}
\mathcal{F}_{V}(\mathbf{Z},\bm{\Theta},\phi(\bflambda)) & =\int_{\bflambda}\phi(\bflambda)\int_{u}p(u|\bflambda)
\left\{ \log p(\bfy|u)+\log\frac{p(\bflambda)}{\phi(\bflambda)}\right\} \dif{u}\,\dif{\bflambda}.
\end{align*}
To compute this bound we first consider the integral 
\[
\log\textrm{T}(\bflambda,\bfy)=\int_{u}p(u|\bflambda)\log p(\bfy|u)\dif{u}.
\]
Since this is simply the expectation of a Gaussian under a Gaussian
we can compute the result analytically as follows%
\begin{align*}
\log\textrm{T}(\bflambda,\bfy) & =\sum_{d=1}^{D}\int_{u}p(u|\bflambda)\left\{ -\frac{N}{2}\log2\pi-\frac{1}{2}\log|
\bm{\Sigma}|-\frac{1}{2}\tr\left[\bm{\Sigma}^{-1}\left(\bfy_{d}\bfy_{d}^{\top}-2\bfy_{d}\bff^{\top}_{d}+
\bff_{d}\bff_{d}^{\top}\right)\right]\right\} \dif{u}.
\end{align*}
 We need to compute $\mathbb{E}_{u|\bflambda}\left[\bff_{d}\right]$
and $\mathbb{E}_{u|\bflambda}\left[\bff_{d}\bff_{d}^{\top}\right]$.
$\mathbb{E}_{u|\bflambda}\left[\bff_{d}\right]$ is a vector with
elements
\begin{align*}
\mathbb{E}_{u|\bflambda}\left[f_{d}(\mathbf{x}_{n})\right]= & \sum_{q=1}^{Q}\int_{\mathcal{Z}}G_{d,q}(\mathbf{x}_{n}-\mathbf{z}')
\mathbb{E}_{u|\bflambda}[u_{q}(\mathbf{z}')]\dif\mathbf{z}'.
\end{align*}
 Assuming that the latent functions are independent GPs, $\mathbb{E}_{u|\bflambda}[u_{q}(\mathbf{z}')]=
\mathbb{E}_{u_{q}|\bflambda_{q}}[u_{q}(\mathbf{z}')]=k_{u_{q}\bflambda_{q}}(\mathbf{z'},\mathbf{Z})
\kMatrix_{\bflambda_{q},\bflambda_{q}}^{-1}(\mathbf{Z},\mathbf{Z})\bflambda_{q}$.
Then %
\begin{align*}
\mathbb{E}_{u|\bflambda}\left[f_{d}(\mathbf{x}_{n})\right] & =\sum_{q=1}^{Q}k_{f_{d}\bflambda_{q}}(\mathbf{x}_{n},
\mathbf{Z})\kMatrix_{\bflambda_{q}\bflambda_{q}}^{-1}(\mathbf{Z},\mathbf{Z})\bflambda_{q}.
\end{align*}
$\mathbb{E}_{u|\bflambda}\left[\bff_{d}\right]$ can be expressed as 
\begin{align*}
\mathbb{E}_{u|\bflambda}\left[\bff_{d}\right]=\kMatrix_{\bff_{d}\bflambda}
\kMatrix_{\bflambda\bflambda}^{-1}\bflambda=\bm{\alpha}_{d}(\mathbf{X},\bflambda)=\bm{\alpha}_{d}.
\end{align*}
On the other hand, $\mathbb{E}_{u|\bflambda}\left[\bff_{d}\bff_{d}^{\top}\right]$
is a matrix with elements 
\begin{align*}
\mathbb{E}_{u|\bflambda}\left[f_{d}(\mathbf{x}_{n})f_{d}(\mathbf{x}_{m})\right]= & \sum_{q=1}^{Q}\int_{\mathcal{Z}}G_{d,q}
(\mathbf{x}_{n}-\mathbf{z})\int_{\mathcal{Z}}G_{d,q}(\mathbf{x}_{m}-\mathbf{z}')\mathbb{E}_{u|\bflambda}[u_{q}(\mathbf{z})
u_{q}(\mathbf{z}')]\dif\mathbf{z}\dif\mathbf{z}'+\alpha_{d}(\mathbf{x}_{n})\alpha_{d}(\mathbf{x}_{m}).
\end{align*}
 With independent GPs the term $\mathbb{E}_{u|\bflambda}[u_{q}(\mathbf{z})u_{q}(\mathbf{z'})]$
can be expressed as 
\[
\mathbb{E}_{u|\bflambda}[u_{q}(\mathbf{z})u_{q}(\mathbf{z}')]=k_{u_{q}u_{q}}(\mathbf{z},\mathbf{z'})-
k_{u_{q}\bflambda_{q}}(\mathbf{z},\mathbf{Z})\kMatrix_{\bflambda_{q}\bflambda_{q}}^{-1}(\mathbf{Z},\mathbf{Z})
k_{u_{q}\bflambda_{q}}^{\top}(\mathbf{z'},\mathbf{Z}).
\]
In this way %
\[
\mathbb{E}_{u|\bflambda}\left[\bff_{d}\bff_{d}^{\top}\right]=\bm{\alpha}_{d}\bm{\alpha}_{d}^{\top}+\kMatrix_{\bff_{d}\bff_{d}}
-\kMatrix_{\bff_{d}\bflambda}\kMatrix_{\bflambda
\bflambda}^{-1}\kMatrix_{\bflambda\bff_{d}}
=\bm{\alpha}_{d}\bm{\alpha}_{d}^{\top}+\widetilde{\kMatrix}_{dd},
\]
with $\widetilde{\kMatrix}_{dd}=\kMatrix_{\bff_{d}\bff_{d}}-\kMatrix_{\bff_{d}\bflambda}\kMatrix_{\bflambda
\bflambda}^{-1}\kMatrix_{\bflambda\bff_{d}}$.

\noindent The expression for $\log\textrm{T}(\bflambda,\bfy)$ is given as
\begin{align*}
\log\textrm{T}(\bflambda,\bfy) & =\log\mathcal{N}\left(\bfy|\bm{\alpha},\bm{\Sigma}\right)-
\frac{1}{2}\sum_{d=1}^{D}\tr\left(\bm{\Sigma}^{-1}\widetilde{\kMatrix}_{dd}\right).
\end{align*}

\noindent The variational lower bound is now given as 
\begin{align}
\mathcal{F}_{V}(\mathbf{Z},\bm{\Theta},\phi) & =\int_{\bflambda}\phi(\bflambda)\log\left\{ \frac{\mathcal{N}\left(\bfy|\bm{\alpha},
\bm{\Sigma}\right)p(\bflambda)}{\phi(\bflambda)}\right\} \dif{\bflambda}-\frac{1}{2}\sum_{d=1}^{D}
\tr\left(\bm{\Sigma}^{-1}\widetilde{\kMatrix}_{dd}\right).\label{eq:var:lower:bound}
\end{align}
 A free form optimization over $\phi(\bflambda)$ could now be performed,
but it is far simpler to reverse Jensen's inequality on the first
term, we then recover the value of the lower bound for optimized $\phi(\bflambda)$
without ever having to explicitly optimise $\phi(\bflambda)$. Reversing
Jensen's inequality, we have%
\[
\mathcal{F}_{V}(\mathbf{Z},\bm{\Theta})=\log\mathcal{N}\left(\bfy|\mathbf{0},\kMatrix_{\bff\bflambda}
\kMatrix_{\bflambda\bflambda}^{-1}\kMatrix_{\bflambda\bff}+\bm{\Sigma}\right)-\frac{1}{2}\sum_{d=1}^{D}
\tr\left(\bm{\Sigma}^{-1}\widetilde{\kMatrix}_{dd}\right).
\]
The form of $\phi(\bflambda)$ which leads to this bound can be found
as 
\begin{align*}
\phi(\bflambda) & \propto\mathcal{N}\left(\bfy|\bm{\alpha},\bm{\Sigma}\right)p(\bflambda)\\
 & =\mathcal{N}\left(\bflambda|\bm{\Sigma}_{\bflambda|\bfy}\kMatrix_{\bflambda\bflambda}^{-1}\kMatrix_{\bflambda\bff}\bm{\Sigma}
^{-1}\bfy,\bm{\Sigma}_{\bflambda|\bfy}\right)
 =\mathcal{N}\left(\kMatrix_{\bflambda\bflambda}\mathbf{A}^{-1}\kMatrix_{\bflambda\bff}
\bm{\Sigma}^{-1}\bfy,\kMatrix_{\bflambda\bflambda}\mathbf{A}^{-1}\kMatrix_{\bflambda\bflambda}\right),
\end{align*}
 with $\bm{\Sigma}_{\bflambda|\bfy}=\left(\kMatrix_{\bflambda\bflambda}^{-1}+\kMatrix_{\bflambda\bflambda}^{-1}\kMatrix_{\bflambda\bff}
\bm{\Sigma}^{-1}\kMatrix_{\bff\bflambda}\kMatrix_{\bflambda\bflambda}^{-1}\right)^{-1}=
\kMatrix_{\bflambda\bflambda}\mathbf{A}^{-1}\kMatrix_{\bflambda\bflambda}$
and $\mathbf{A}=\kMatrix_{\bflambda\bflambda}+\kMatrix_{\bflambda\bff}\bm{\Sigma}^{-1}\kMatrix_{\bff\bflambda}$.

\subsection{Predictive distribution}

The predictive distribution for a new test point given the training
data is also required. This can be expressed as 
\begin{align*}
p\left(\bfy_{*}|\bfy,\mathbf{X},\mathbf{Z}\right) & =\int_{u,\bflambda}p(\bfy_{*}|u)q(u,\bflambda)\dif\bflambda\dif{u}
=\int_{u,\bflambda}p(\bfy_{*}|u)p(u|\bflambda)\phi(\bflambda)\dif\bflambda\dif{u}\\
&=\int_{u}p(\bfy_{*}|u)\left[\int_{\bflambda}p(u|\bflambda)\phi(\bflambda)\dif\bflambda\right]\dif{u}.
\end{align*}
Using the Gaussian form for the $\phi(\bflambda)$ we can compute
\begin{align*}
\int_{\bflambda}p(u|\bflambda)\phi(\bflambda)\dif\bflambda & =\int_{\bflambda}\mathcal{N}(u|k_{u\bflambda}
\kMatrix_{\bflambda\bflambda}^{-1}\bflambda,k_{uu}-k_{u\bflambda}\kMatrix_{\bflambda\bflambda}^{-1}k_{\bflambda u})\\
&\times\mathcal{N}\left(\kMatrix_{\bflambda\bflambda}\mathbf{A}^{-1}\kMatrix_{\bflambda\bff}\bm{\Sigma}^{-1}\bfy,
\kMatrix_{\bflambda\bflambda}\mathbf{A}^{-1}\kMatrix_{\bflambda\bflambda}\right)\dif\bflambda\\
 & =\mathcal{N}\left(u|k_{u\bflambda}\mathbf{A}^{-1}\kMatrix_{\bflambda\bff}\bm{\Sigma}^{-1}\bfy,k_{uu}-k_{u\bflambda}\left(\kMatrix_{\bflambda\bflambda}^{-1}-\mathbf{A}^{-1}\right)k_{\bflambda u}\right).
\end{align*}
 Which allows us to write the predictive distribution as 
\begin{align*}
p\left(\bfy_{*}|\bfy,\mathbf{X},\mathbf{Z}\right) & =\int_{u}\mathcal{N}\left(\bfy_{*}|\bff_{*},
\bm{\Sigma}_*\right)\mathcal{N}\left(u|\bm{\mu}_{u|\bflambda},\bm{\Sigma}_{u|\bflambda}\right)\dif{u}
=\mathcal{N}\left(\bfy_{*}|\bm{\mu}_{\bfy_{*}},\bm{\Sigma}_{\bfy_{*}}\right)
\end{align*}
with $\bm{\mu}_{\bfy_{*}}=\kMatrix_{\bff_{*}\bflambda}\mathbf{A}^{-1}\kMatrix_{\bflambda\bff}\bm{\Sigma}^{-1}\bfy$
and $\bm{\Sigma}_{\bfy_{*}}=\kMatrix_{\bff_{*}\bff_{*}}-\kMatrix_{\bff_{*}\bflambda}\left(\kMatrix_{\bflambda\bflambda}^{-1}-
\mathbf{A}^{-1}\right)\kMatrix_{\bflambda\bff_{*}}+\bm{\Sigma}_{*}$.

\subsection{Optimisation of the Bound}

Optimisation of the bound (with respect to the variational parameters
and the parameters of the covariance functions) can be carried out
through gradient based methods. We follow the notation of \citet{Brookes:matrix05}
obtaining similar results to \citet{Lawrence:larger07}. This notation
allows us to apply the chain rule for matrix derivation in a straight-forward
manner. The resulting gradients can then be combined with gradients
of the covariance functions with respect to their parameters to optimize
the model.

Let's define $\mathbf{G}\veC=\vecO\mathbf{G}$, where $\vecO$ is
the vectorization operator over the matrix $\mathbf{G}$. For a function
$\mathcal{F}_{V}(\mathbf{Z})$ the equivalence between $\frac{\partial \mathcal{F}_{V}(\mathbf{Z})}{\partial\mathbf{G}}$
and $\frac{\partial \mathcal{F}_{V}(\mathbf{Z})}{\partial\mathbf{G}\veC}$ is
given through $\frac{\partial \mathcal{F}_{V}(\mathbf{Z})}{\partial\mathbf{G}\veC}=\left(\left(\frac{\partial \mathcal{F}_{V}(\mathbf{Z})}{\partial\mathbf{G}}\right)\veC\right)^{\top}$.
The log-likelihood function is given as 
\begin{align*}
\mathcal{F}_{V}(\mathbf{Z},\bm{\Theta})\propto -\frac{1}{2}\log\abs{\bm{\Sigma}+\kMatrix_{\bff\bflambda}\kMatrix_{\mathbf{\bflambda\bflambda}}^{-1}\kMatrix_{\bflambda\bff}}-\frac{1}{2}\tr\left[\left(\bm{\Sigma}+\kMatrix_{\bff\bflambda}\kMatrix_{\mathbf{\bflambda\bflambda}}^{-1}\kMatrix_{\bflambda\bff}\right)^{-1}\bfy\bfy^{\top}\right]-\frac{1}{2}\tr\left(\bm{\Sigma}^{-1}\widetilde{\kMatrix}\right),
\end{align*}
 where $\widetilde{\kMatrix}=\kMatrix_{\bff\bff}-\kMatrix_{\bff\bflambda}\kMatrix_{\bflambda\bflambda}^{-1}\kMatrix_{\bflambda\bff}$.
Using the matrix inversion lemma and its equivalent form for determinants,
the above expression can be written as 
\begin{align*}
\mathcal{F}_{V}(\mathbf{Z},\bm{\Theta})\propto & \frac{1}{2}\log\abs{\kMatrix_{\bflambda\bflambda}}-\frac{1}{2}\log\abs{\mathbf{A}}-\frac{1}{2}\log\abs{\bm{\Sigma}}-\frac{1}{2}\tr\left[\bm{\Sigma}^{-1}\bfy\bfy^{\top}\right]\\
&+\frac{1}{2}\tr\left[\bm{\Sigma}^{-1}\kMatrix_{\bff\bflambda}\mathbf{A}^{-1}
\kMatrix_{\bflambda\bff}\bm{\Sigma}^{-1}\bfy\bfy^{\top}\right]
-\frac{1}{2}\tr\left(\bm{\Sigma}^{-1}\widetilde{\kMatrix}\right),
\end{align*}
 up to a constant. We can find $\frac{\partial \mathcal{F}_{V}(\mathbf{Z})}{\partial\bm{\theta}}$
and $\frac{\partial \mathcal{F}_{V}(\mathbf{Z})}{\partial\mathbf{Z}}$ applying
the chain rule to $\mathcal{F}_{V}(\mathbf{Z},\bm{\Theta})$ obtaining expressions for $\frac{\partial \mathcal{F}_{V}(\mathbf{Z})}{\partial\kMatrix_{\bff\bff}}$,
$\frac{\partial \mathcal{F}_{V}(\mathbf{Z})}{\partial\kMatrix_{\bff\bflambda}}$
and $\frac{\partial \mathcal{F}_{V}(\mathbf{Z})}{\partial\kMatrix_{\bflambda\bflambda}}$
and combining those with the relevant derivatives of the covariances
wrt $\bm{\Theta}$, $\mathbf{Z}$ and the parameters associated to the model kernels,   
\begin{align}
\frac{\partial\mathcal{F}}{\partial\mathbf{G}\veC}=\left[\frac{\partial\mathcal{F}_{\mathbf{A}}}{\partial\mathbf{A}\veC}\frac{\partial\mathbf{A}\veC}{\partial\mathbf{G}\veC}\right]\delta_{GK}+\frac{\partial\mathcal{F}_{\mathbf{G}}}{\partial\mathbf{G}\veC},\label{eq:chain:rule}
\end{align}
 where the subindex in $\mathcal{F}_{\mathbf{E}}$ stands for those
terms of $\mathcal{F}$ which depend on $\mathbf{E}$, $\mathbf{G}$
is either $\kMatrix_{\bff\bff}$, $\kMatrix_{\bflambda\bff}$
or $\kMatrix_{\bflambda\bflambda}$ and $\delta_{GK}$ is zero if
$\mathbf{G}$ is equal to $\kMatrix_{\bff\bff}$ and one in other
case. For convenience we have used $\mathcal{F}\equiv \mathcal{F}_{V}(\mathbf{Z},\bm{\Theta})$.
Next we present expressions for each partial derivative 
\begin{gather*}
\begin{split}
\frac{\partial\mathbf{A}\veC}{\partial\bm{\Sigma}\veC} & =-\left(\kMatrix_{\bflambda,\bff}
\bm{\Sigma}^{-1}\otimes\kMatrix_{\bflambda,\bff}\bm{\Sigma}^{-1}\right),
\quad\frac{\partial\mathcal{F}_{\bm{\Sigma}}}{\partial\bm{\Sigma}\veC}=
-\frac{1}{2}\left(\left(\bm{\Sigma}^{-1}\mathbf{H}\bm{\Sigma}^{-1}\right)\veC\right)^{\top}+
\frac{1}{2}\left(\left(\bm{\Sigma}^{-1}\widetilde{\kMatrix}^{\top}\bm{\Sigma}^{-1}\right)
\veC\right)^{\top}\label{eq:grads:def:CH}
\end{split}
\\
\begin{split}\frac{\partial\mathbf{A}\veC}{\partial\kMatrix_{\bflambda,\bflambda}\veC}= & \;\mathbf{I},\quad\frac{\partial\mathbf{A}\veC}{\partial\kMatrix_{\bflambda,\bff}\veC}=\left(\kMatrix_{\bflambda,\bff}\bm{\Sigma}^{-1}\otimes\mathbf{I}\right)+\left(\mathbf{I}\otimes\kMatrix_{\bflambda,\bff}\bm{\Sigma}^{-1}\right)\mathbf{T_{A}}\label{eq:deriv:TA},\quad\frac{\partial\mathcal{F}_{\kMatrix_{\bff,\bff}}}{\partial\kMatrix_{\bff,\bff}\veC}=-\frac{1}{2}\bm{\Sigma}^{-1}\veC\end{split}
\\
\begin{split}
\frac{\partial\mathcal{F}_{\mathbf{A}}}{\partial\mathbf{A}\veC}=-\frac{1}{2}\left(\mathbf{C}\veC\right)^{\top},\quad\frac{\partial\mathcal{F}_{\kMatrix_{\bflambda,\bff}}}{\partial\kMatrix_{\bflambda,\bff}\veC}=\left(\left(\mathbf{A}^{-1}\kMatrix_{\bflambda,\bff}\bm{\Sigma}^{-1}\bfy\bfy^{\top}\bm{\Sigma}^{-1}\right)\veC\right)^{\top}+\left(\left(\kMatrix_{\bflambda,\bflambda}^{-1}\kMatrix_{\bflambda,\bff}\bm{\Sigma}^{-1}\right)\veC\right)^{\top},\end{split}
\\
\begin{split}\frac{\partial\mathcal{F}_{\kMatrix_{\bflambda,\bflambda}}}{\partial\kMatrix_{\bflambda,\bflambda}\veC}=\frac{1}{2}\left(\left(\kMatrix_{\bflambda,\bflambda}^{-1}\right)\veC\right)^{\top}-\frac{1}{2}\left(\left(\kMatrix_{\bflambda,\bflambda}^{-1}\kMatrix_{\bflambda,\bff}\bm{\Sigma}^{-1}\kMatrix_{\bff,\bflambda}\kMatrix_{\bflambda,\bflambda}^{-1}\right)\veC\right)^{\top},
\end{split}
\end{gather*}
 where $\mathbf{C}=\mathbf{A}^{-1}+\mathbf{A}^{-1}\kMatrix_{\bflambda,\bff}\bm{\Sigma}^{-1}\bfy\bfy^{\top}\bm{\Sigma}^{-1}\kMatrix_{\bff,\bflambda}\mathbf{A}^{-1}$,
$\mathbf{H}=\bm{\Sigma}-\bfy\bfy^{\top}+\kMatrix_{\bff,\bflambda}\mathbf{A}^{-1}\kMatrix_{\bflambda,\bff}\bm{\Sigma}^{-1}\bfy\bfy^{\top}+\left(\kMatrix_{\bff,\bflambda}\mathbf{A}^{-1}\kMatrix_{\bflambda,\bff}\bm{\Sigma}^{-1}\bfy\bfy^{\top}\right)^{\top}$
and $\mathbf{T_{A}}$ is a \emph{vectorized transpose matrix} \citep{Brookes:matrix05}
and we have not included its dimensions to keep the notation clearer.
We can replace the above expressions in \eqref{eq:chain:rule} to
find the corresponding derivatives, so \begin{align*}
\frac{\partial\mathcal{F}}{\partial\kMatrix_{\bff,\bff}\veC}=-\frac{1}{2}\bm{\Sigma}^{-1}\veC\end{align*}
 We also have 
\begin{align*}
\begin{split}\frac{\partial\mathcal{F}}{\partial\kMatrix_{\bflambda,\bff}\veC}= & -\frac{1}{2}\left(\mathbf{C}\veC\right)^{\top}\left[\left(\kMatrix_{\bflambda,\bff}\bm{\Sigma}^{-1}\otimes\mathbf{I}\right)+\left(\mathbf{I}\otimes\kMatrix_{\bflambda,\bff}\bm{\Sigma}^{-1}\right)\mathbf{T_{A}}\right]+\left(\left(\mathbf{A}^{-1}\kMatrix_{\bflambda,\bff}\bm{\Sigma}^{-1}\bfy\bfy^{\top}\bm{\Sigma}^{-1}\right)\veC\right)^{\top}\\
&+\left(\left(\kMatrix_{\bflambda,\bflambda}^{-1}\kMatrix_{\bflambda,\bff}\bm{\Sigma}^{-1}\right)\veC\right)^{\top}\\
= & \left(\left(-\mathbf{C}\kMatrix_{\bflambda,\bff}\bm{\Sigma}^{-1}+\mathbf{A}^{-1}\kMatrix_{\bflambda,\bff}\bm{\Sigma}^{-1}\bfy\bfy^{\top}\bm{\Sigma}^{-1}+\kMatrix_{\bflambda,\bflambda}^{-1}\kMatrix_{\bflambda,\bff}\bm{\Sigma}^{-1}\right)\veC\right)^{\top}.
\end{split}
\end{align*}
Finally, results for $\frac{\partial\mathcal{F}}{\partial\kMatrix_{\bflambda,\bff}\veC}$
and $\frac{\partial\mathcal{F}}{\partial\bm{\Sigma}\veC}$
are obtained as \begin{align*}
\frac{\partial\mathcal{F}}{\partial\kMatrix_{\bflambda,\bflambda}\veC}= & -\frac{1}{2}\left(\mathbf{C}\veC\right)^{\top}+\frac{1}{2}\left(\left(\kMatrix_{\bflambda,\bflambda}^{-1}\right)\veC\right)^{\top}-\frac{1}{2}\left(\left(\kMatrix_{\bflambda,\bflambda}^{-1}\kMatrix_{\bflambda,\bff}\bm{\Sigma}^{-1}\kMatrix_{\bff,\bflambda}\kMatrix_{\bflambda,\bflambda}^{-1}\right)\veC\right)^{\top}\\
\frac{\partial\mathcal{F}}{\partial\bm{\Sigma}\veC}= & \frac{1}{2}\left(\left(\bm{\Sigma}^{-1}\left(\widetilde{\kMatrix}^{\top}-\mathbf{H}\right)\bm{\Sigma}^{-1}\right)\veC\right)^{\top}+\frac{1}{2}\left(\mathbf{C}\veC\right)^{\top}\left(\kMatrix_{\bflambda,\bff}\bm{\Sigma}^{-1}\otimes\kMatrix_{\bflambda,\bff}\bm{\Sigma}^{-1}\right).
\end{align*}

\bibliography{../sparseMultigp/reportbiblio}

\begin{thebibliography}{24}
\providecommand{\natexlab}[1]{#1}
\providecommand{\url}[1]{\texttt{#1}}
\expandafter\ifx\csname urlstyle\endcsname\relax
  \providecommand{\doi}[1]{doi: #1}\else
  \providecommand{\doi}{doi: \begingroup \urlstyle{rm}\Url}\fi

\bibitem[\'{A}lvarez and Lawrence(2009)]{Alvarez:sparse2009}
Mauricio \'{A}lvarez and Neil~D. Lawrence.
\newblock Sparse convolved {G}aussian processes for multi-output regression.
\newblock In \emph{NIPS}, volume~21, pages 57--64. MIT Press, Cambridge, MA,
  2009.

\bibitem[\'{A}lvarez et~al.(2009)\'{A}lvarez, Luengo, and
  Lawrence]{Alvarez:lfm09}
Mauricio \'{A}lvarez, David Luengo, and Neil~D. Lawrence.
\newblock {L}atent {F}orce {M}odels.
\newblock In  \citet{vanDyk:aistats09}, pages 9--16.

\bibitem[Bishop(2006)]{Bishop:PRLM06}
Christopher~M. Bishop.
\newblock \emph{Pattern Recognition and Machine Learning}.
\newblock Information Science and Statistics. Springer, 2006.

\bibitem[Bonilla et~al.(2008)Bonilla, Chai, and Williams]{Bonilla:multi07}
Edwin~V. Bonilla, Kian~Ming Chai, and Christopher K.~I. Williams.
\newblock Multi-task {G}aussian process prediction.
\newblock In John~C. Platt, Daphne Koller, Yoram Singer, and Sam Roweis,
  editors, \emph{NIPS}, volume~20, Cambridge, MA, 2008. MIT Press.

\bibitem[Boyle and Frean(2005)]{Boyle:dependent04}
Phillip Boyle and Marcus Frean.
\newblock Dependent {G}aussian processes.
\newblock In Lawrence Saul, Yair Weiss, and L\'eon Bouttou, editors,
  \emph{NIPS}, volume~17, pages 217--224, Cambridge, MA, 2005. MIT Press.

\bibitem[Brookes(2005)]{Brookes:matrix05}
Michael Brookes.
\newblock The matrix reference manual.
\newblock Available on-line., 2005.
\newblock \url{http://www.ee.ic.ac.uk/hp/staff/dmb/matrix/intro.html}.

\bibitem[Csat\'o and Opper(2001)]{Csato:sparse00}
Lehel Csat\'o and Manfred Opper.
\newblock Sparse representation for {G}aussian process models.
\newblock In Todd~K. Leen, Thomas~G. Dietterich, and Volker Tresp, editors,
  \emph{NIPS}, volume~13, pages 444--450, Cambridge, MA, 2001. MIT Press.

\bibitem[Evgeniou et~al.(2005)Evgeniou, Micchelli, and
  Pontil]{Evgeniou:multitask05}
Theodoros Evgeniou, Charles~A. Micchelli, and Massimiliano Pontil.
\newblock Learning multiple tasks with kernel methods.
\newblock \emph{Journal of Machine Learning Research}, 6:\penalty0 615--637,
  2005.

\bibitem[Gao et~al.(2008)Gao, Honkela, Rattray, and Lawrence]{Gao:latent08}
Pei Gao, Antti Honkela, Magnus Rattray, and Neil~D. Lawrence.
\newblock Gaussian process modelling of latent chemical species: Applications
  to inferring transcription factor activities.
\newblock \emph{Bioinformatics}, 24:\penalty0 i70--i75, 2008.
\newblock \doi{10.1093/bioinformatics/btn278}.

\bibitem[Goovaerts(1997)]{Goovaerts:book97}
Pierre Goovaerts.
\newblock \emph{{G}eostatistics {F}or {N}atural {R}esources {E}valuation}.
\newblock Oxford University Press, USA, 1997.

\bibitem[Higdon(2002)]{Higdon:convolutions02}
David~M. Higdon.
\newblock Space and space-time modelling using process convolutions.
\newblock In C.~Anderson, V.~Barnett, P.~Chatwin, and A.~El-Shaarawi, editors,
  \emph{Quantitative methods for current environmental issues}, pages 37--56.
  Springer-Verlag, 2002.

\bibitem[Journel and Huijbregts(1978)]{Journel:miningBook78}
Andre~G. Journel and Charles~J. Huijbregts.
\newblock \emph{Mining Geostatistics}.
\newblock Academic Press, London, 1978.
\newblock ISBN 0-12391-050-1.

\bibitem[Lawrence(2007)]{Lawrence:larger07}
Neil~D. Lawrence.
\newblock Learning for larger datasets with the {G}aussian process latent
  variable model.
\newblock In Marina Meila and Xiaotong Shen, editors, \emph{AISTATS 11}, San
  Juan, Puerto Rico, 21-24 March 2007. Omnipress.

\bibitem[Lawrence et~al.(2003)Lawrence, Seeger, and Herbrich]{Lawrence:ivm02}
Neil~D. Lawrence, Matthias Seeger, and Ralf Herbrich.
\newblock Fast sparse {G}aussian process methods: The informative vector
  machine.
\newblock In Sue Becker, Sebastian Thrun, and Klaus Obermayer, editors,
  \emph{NIPS}, volume~15, pages 625--632, Cambridge, MA, 2003. MIT Press.

\bibitem[Lawrence et~al.(2007)Lawrence, Sanguinetti, and
  Rattray]{Lawrence:gpsim2007a}
Neil~D. Lawrence, Guido Sanguinetti, and Magnus Rattray.
\newblock Modelling transcriptional regulation using {G}aussian processes.
\newblock In Bernhard Sch\"olkopf, John~C. Platt, and Thomas Hofmann, editors,
  \emph{NIPS}, volume~19, pages 785--792. MIT Press, Cambridge, MA, 2007.

\bibitem[L\'{a}zaro-Gredilla and
  Figueiras-Vidal(2010)]{Lazaro:interdomain:2010}
Miguel L\'{a}zaro-Gredilla and An\'{i}bal Figueiras-Vidal.
\newblock Inter-domain {G}aussian processes for sparse inference using inducing
  features.
\newblock In \emph{NIPS}, volume~22, pages 1087--1095. MIT Press, Cambridge,
  MA, 2010.

\bibitem[Osborne et~al.(2008)Osborne, Rogers, Ramchurn, Roberts, and
  Jennings]{Rogers:towards08}
Michael~A. Osborne, Alex Rogers, Sarvapali~D. Ramchurn, Stephen~J. Roberts, and
  Nicholas~R. Jennings.
\newblock Towards real-time information processing of sensor network data using
  computationally efficient multi-output {G}aussian processes.
\newblock In \emph{Proceedings of the International Conference on Information
  Processing in Sensor Networks (IPSN 2008)}, 2008.

\bibitem[{Qui\~nonero Candela} and Rasmussen(2005)]{Quinonero:unifying05}
Joaquin {Qui\~nonero Candela} and Carl~Edward Rasmussen.
\newblock A unifying view of sparse approximate {G}aussian process regression.
\newblock \emph{Journal of Machine Learning Research}, 6:\penalty0 1939--1959,
  2005.

\bibitem[Rasmussen and Williams(2006)]{Rasmussen:book06}
Carl~Edward Rasmussen and Christopher K.~I. Williams.
\newblock \emph{{G}aussian Processes for Machine Learning}.
\newblock MIT Press, Cambridge, MA, 2006.
\newblock ISBN 0-262-18253-X.

\bibitem[Seeger et~al.(2003)Seeger, Williams, and Lawrence]{Seeger:fast03}
Matthias Seeger, Christopher K.~I. Williams, and Neil~D. Lawrence.
\newblock Fast forward selection to speed up sparse {G}aussian process
  regression.
\newblock In Christopher~M. Bishop and Brendan~J. Frey, editors,
  \emph{Proceedings of the Ninth International Workshop on Artificial
  Intelligence and Statistics}, Key West, FL, 3--6 Jan 2003.

\bibitem[Snelson and Ghahramani(2006)]{Snelson:pseudo05}
Edward Snelson and Zoubin Ghahramani.
\newblock Sparse {G}aussian processes using pseudo-inputs.
\newblock In Yair Weiss, Bernhard Sch\"olkopf, and John~C. Platt, editors,
  \emph{NIPS}, volume~18, Cambridge, MA, 2006. MIT Press.

\bibitem[Teh et~al.(2005)Teh, Seeger, and Jordan]{Teh:semiparametric05}
Yee~Whye Teh, Matthias Seeger, and Michael~I. Jordan.
\newblock Semiparametric latent factor models.
\newblock In Robert~G. Cowell and Zoubin Ghahramani, editors, \emph{AISTATS
  10}, pages 333--340, Barbados, 6-8 January 2005. Society for Artificial
  Intelligence and Statistics.

\bibitem[Titsias(2009)]{Titsias:variational09}
Michalis~K. Titsias.
\newblock Variational learning of inducing variables in sparse {G}aussian
  processes.
\newblock In  \citet{vanDyk:aistats09}, pages 567--574.

\bibitem[van Dyk and Welling(2009)]{vanDyk:aistats09}
David van Dyk and Max Welling, editors.
\newblock \emph{AISTATS}, Clearwater Beach, Florida, 16-18 April 2009. JMLR
  W\&CP 5.

\end{thebibliography}

\end{document}